\documentclass[10pt,twocolumn,letterpaper]{article}
\usepackage[pagenumbers]{cvpr} 
\usepackage{graphicx}
\usepackage{amsmath}
\usepackage{amssymb}
\usepackage{booktabs}
\usepackage{tabularx}
\usepackage{float}
\usepackage{multirow}
\usepackage{siunitx}
\usepackage{hyperref}
\hypersetup{colorlinks,allcolors=black}
\usepackage{caption}  
\usepackage[capitalize]{cleveref}
\usepackage{pifont}
\newcommand{\xmark}{\ding{55}}%

\crefname{section}{Sec.}{Secs.}
\Crefname{section}{Section}{Sections}
\Crefname{table}{Table}{Tables}
\crefname{table}{Tab.}{Tabs.}

\begin{document}

\title{\ MAIR: Multi-view Attention Inverse Rendering with 3D Spatially-Varying Lighting Estimation}

\author{
{JunYong Choi$^{1,2}$}\quad 
{SeokYeong Lee$^{1,2}$}\quad
{Haesol Park$^{1}$}\quad 
{Seung-Won Jung$^{2}$}\quad
{Ig-Jae Kim$^{1,3,4}$}\quad
{Junghyun Cho$^{1,3,4}$} 
\\[2mm]
{$^{1}$Korea Institute of Science and Technology(KIST)}\quad 
{$^{2}$Korea University} \\
{$^{3}$AI-Robotics, KIST School, University of Science and Technology} \\
{$^{4}$Yonsei-KIST Convergence Research Institute, Yonsei University} \\
\vspace{-3mm}
{\tt\small \{happily,shapin94,haesol,drjay,jhcho\}@kist.re.kr}\quad
{\tt\small swjung83@korea.ac.kr}
\renewcommand\footnotemark{}
\renewcommand\footnoterule{}
\thanks{This work was partly supported by Institute of Information \& communications Technology Planning \& Evaluation (IITP) grant funded by the Korea government(MSIT)(No.2020-0-00457, 50\%) and KIST Institutional Program(Project No.2E32301, 50\%).}
}
\maketitle

\begin{abstract}
We propose a scene-level inverse rendering framework that uses multi-view images to decompose the scene into geometry, a SVBRDF, and 3D spatially-varying lighting. Because multi-view images provide a variety of information about the scene, multi-view images in object-level inverse rendering have been taken for granted. However, owing to the absence of multi-view HDR synthetic dataset, scene-level inverse rendering has mainly been studied using single-view image. We were able to successfully perform scene-level inverse rendering using multi-view images by expanding OpenRooms dataset and designing efficient pipelines to handle multi-view images, and splitting spatially-varying lighting. Our experiments show that the proposed method not only achieves better performance than single-view-based methods, but also achieves robust performance on unseen real-world scene. Also, our sophisticated 3D spatially-varying lighting volume allows for photorealistic object insertion in any 3D location.

\end{abstract}


\section{Introduction}
\label{sec:intro}
Inverse rendering is a technology used to estimate material, lighting, and geometry from RGB color images. Decomposing a scene through inverse rendering enables various applications such as object insertion, relighting, and material editing in VR and AR. However, since inverse rendering is an ill-posed problem, previous studies have focused only on a part of the inverse rendering, such as intrinsic image decomposition~\cite{id1,id2,id3,id4}, shape from shading~\cite{sfs1,sfs2,sfs3}, and material estimation~\cite{sime1, sime2, sime3, sime4, sime5}. 

\begin{figure}[t!]
  \centering
  \includegraphics[width=\linewidth]{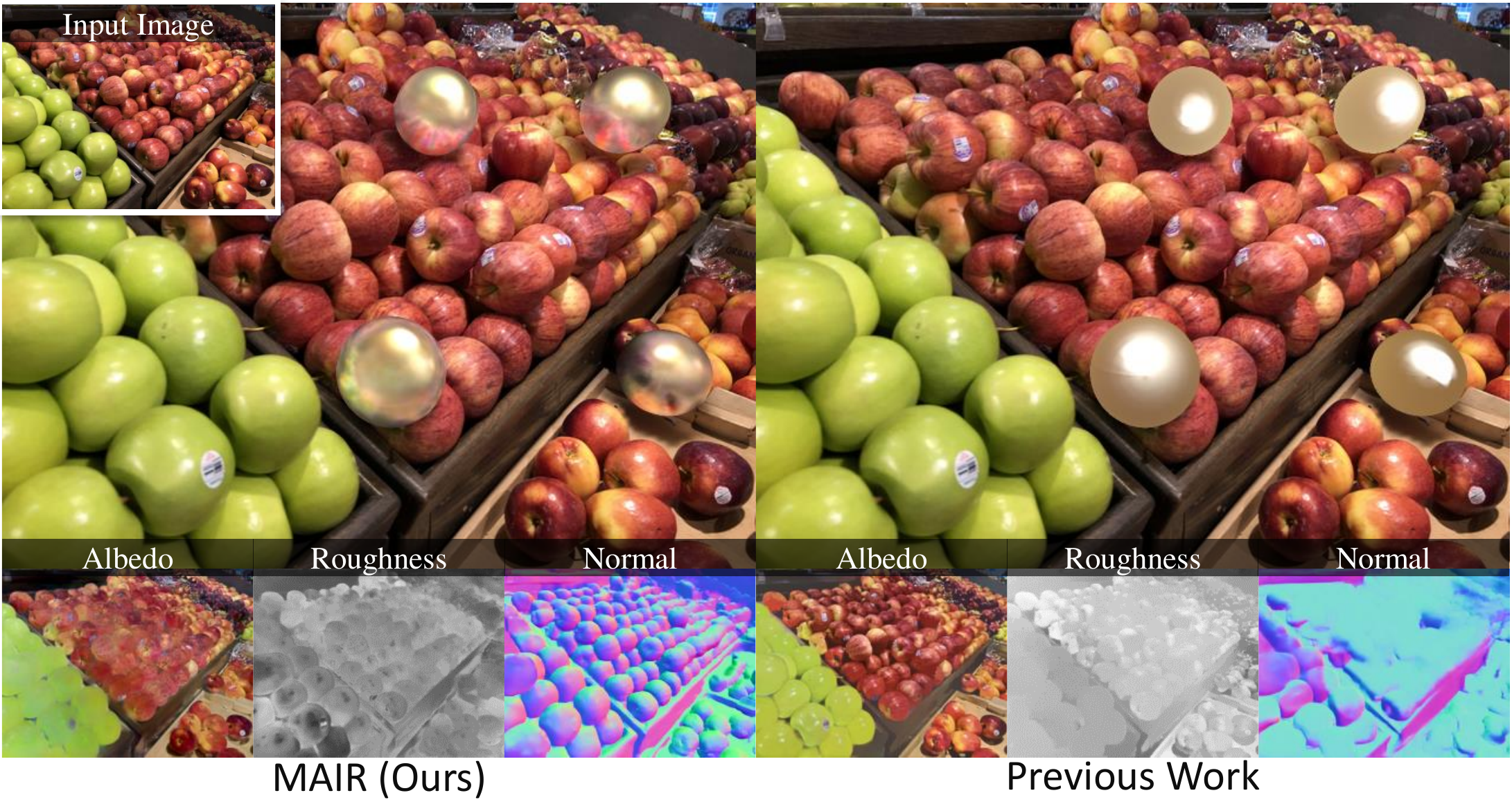}
  \caption{The result of inverse rendering and floating chrome sphere insertion in the unseen real-world scene. Since the single-view-based method\cite{cis2020} relies only on contextual information, it has difficulty estimating the complex material, geometry of the real-world. Notice the spatially-consistent albedo of apples, eloborated normal, and realistic lighting reflected on the inserted object. In previous work\cite{cis2020}, they have limitations on floating object insertion because they use per-pixel lighting.}
  \label{fig:main}
  \vspace{-2mm}
\end{figure}

Recent advances in GPU-accelerated physically-based rendering algorithms have made constructing large-scale photorealistic indoor high dynamic range (HDR) image dataset that include geometries, materials, and spatially-varying lighting~\cite{openrooms2021}. The availability of such dataset and the recent success of deep learning technology have enabled seminal works on single-view-based inverse rendering~\cite{cis2020, irisformer2022, vsg, phyir, Li22, zhu2022montecarlo}. These methods have fundamental limitations in that they are prone to bias in training dataset despite having shown promising results. Specifically, single-view-based inverse rendering must refer to specular reflectance from the contextual information of the image, making it less reliable for predicting complex SVBRDF(spatially-varying bidirectional reflectance distribution function) in the real-world. Fig.~\ref{fig:main} shows such an example, where decomposition has severely failed owing to the complexity of the real-world scene. In addition, depth-scale ambiguity makes it challenging to employ these methods to 3D applications such as object insertion. 

\begin{figure*}[t!]
  \centering
  \includegraphics[width=\linewidth]{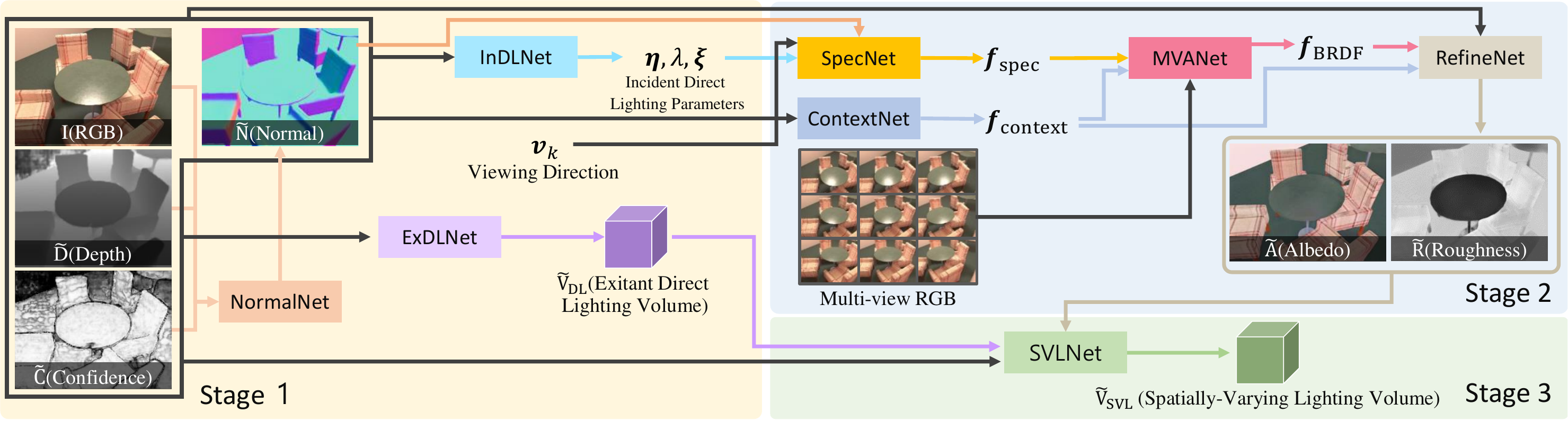}
   \caption{MAIR's entire pipeline. Our method has reduced the difficulty of inverse rendering by splitting the scene components as small as possible, and progressively estimating the scene components.}
   \label{fig:whole}
\end{figure*}

In this paper, we introduce \textbf{MAIR}, the scene-level \textbf{M}ulti-view \textbf{A}ttention \textbf{I}nverse \textbf{R}endering pipeline. MAIR exploits multiple RGB observations of the same scene point from multiple cameras and, more importantly, it utilizes multi-view stereo (MVS) depth as well as scene context to estimate SVBRDF. As a result, the method becomes less dependent on the implicit scene context, and shows better performance on unseen real-world images. However, the processing of multi-view images inherently requires a high computational cost to handle multiple observations with occlusions and brightness mismatches. To remedy this, we design a three-stage training pipeline for MAIR that can significantly increase training efficiency and reduce memory consumption. Spatially-varying lighting consists of direct lighting and indirect lighting. Indirect lighting affected by the surrounding environment makes inverse rendering difficult. Therefore, in Stage 1, we first estimate the direct lighting and geometry, which reflect the amount of light entering each point and in which direction the specular reflection appears. We estimate the material in Stage 2 using the estimates of the direct lighting, geometry, and multi-view color images. In Stage 3, we collect all the material, geometry, and direct lighting information and finally estimate 3D spatially-varying lighting, including indirect lighting. The MAIR pipeline is shown in Fig.~\ref{fig:whole}. We created the OpenRooms Forward Facing(OpenRooms FF) dataset as an extension of OpenRooms~\cite{openrooms2021} to train the proposed network.

Our contribution can be summarized as follows:
\begin{itemize}
\item As summarized in Tab.~\ref{tab:compare_prior}, we believe this is the first demonstration of using multi-view images to decompose the scene into geometry, complex material, and 3D spatially-varying lighting without test-time optimization. Also, we release OpenRooms FF dataset.

\item We propose a framework that can efficiently train multi-view inverse rendering networks. Our framework increases the training efficiency by decomposing lighting and separating the scene components by stage.
\item Our method achieves better inverse rendering performance than the existing single-view-based method, and realistic object insertion in real-world is possible by reproducing 3D lighting of the real-world.
\vspace{-2mm}
\end{itemize}

\section{Related Works}
\noindent\textbf{Inverse rendering.}
Research on inverse rendering has received significant attention in recent years owing to the development of deep learning technology. Yu \etal\cite{outdoorinv} performed outdoor inverse rendering with multi-view self-supervision, but their lighting is simple distant lighting. A pioneering work by Li \etal \cite{cis2020} conducted inverse rendering on a single image, and IRISformer\cite{irisformer2022} further improved the performance by replacing convolutional neural networks (CNNs) with a Transformer\cite{vit}. PhyIR \cite{phyir} addressed this problem using panoramic image. However, the lighting representation in~\cite{cis2020,irisformer2022, phyir} is a 2D per-pixel environment map, which is insufficient for modeling 3D lighting. Li \etal \cite{Li22} adopted a parametric 3D lighting representation; however, it fixed with two types of indoor light sources. The recently introduced Zhu \etal\cite{zhu2022montecarlo} demonstrated realistic 3D volumetric lighting based on ray tracing. But, since these prior works\cite{outdoorinv, cis2020,irisformer2022,Li22, phyir, zhu2022montecarlo} are all single-view-based, they inherently rely on the scene context, making these methods less reliable for unseen images. In contrast, the proposed method can estimate BRDF and geometry more accurately by utilizing the multi-view correspondences as additional cues for inverse rendering. 

\begin{table}[t!]
\footnotesize
\centering
\resizebox{\linewidth}{!}{%
\begin{tabular}{|l|c|c|c|c|}
\hline
Method & \multicolumn{1}{c|}{Input} & \multicolumn{1}{c|}{Material} & \multicolumn{1}{c|}{Lighting} & \multicolumn{1}{c|}{Insertion} \\
\hline
VSG\cite{vsg} &single &diffuse &volume &any\\
Li~\etal\cite{cis2020} &single &microfacet &per-pixel &surface\\
IRISformer\cite{irisformer2022} &single  &microfacet &per-pixel &surface\\
PBE\cite{Li22} &single &microfacet &parametric &any\\
SOLD \cite{SOLD-Net} & single &\xmark &per-pixel &surface\\
Zhu~\etal\cite{zhu2022montecarlo} &single &microfacet(metalic) &volume &any\\
lighthouse~\cite{lighthouse} &stereo &\xmark &volume &any\\
FreeView~\cite{philip2021} &multi &glossy &irradiance & \xmark\\
Zhang~\etal \cite{invrender2022} &multi &microfacet &object & \xmark\\
PhotoScene~\cite{photoscene} &any &microfacet &parameteric &any\\
Intrinsic3D~\cite{maier2017intrinsic3d} &multi RGBD &diffuse &volume &any\\
Zhang \etal\cite{zhang2016emptying} &multi RGBD &diffuse &parameteric &any\\
\hline
MAIR (Ours) &multi &microfacet &volume &any\\
\hline
\end{tabular}%
}
\caption{Compared to previous works, our work is the first demonstration to perform multi-view scene-level inverse rendering.}
\vspace{-3mm}
\label{tab:compare_prior}
\end{table}

\noindent\textbf{Lighting estimation.}
Lighting estimation has been studied not only as a sub-task of the inverse rendering but also an important research topic\cite{le1,le2,le3,Song19}. In Lighthouse \cite{lighthouse}, 3D spatially-varying lighting was obtained by estimating the RGBA lighting volume. Wang \etal\cite{vsg} estimated a more sophisticated 3D lighting volume by replacing RGB with a spherical Gaussian in lighting volume. However, because the methods in~\cite{lighthouse,vsg} assume Lambertian reflectance, they cannot represent complex indirect lighting and have limitations in expressing HDR lighting due to weak-supervision with LDR dataset\cite{interiornet}. On the other hand, the proposed method can handle complex SVBRDF and HDR lighting well because we trained our model on the large indoor HDR dataset\cite{openrooms2021}. Recently, a study on spatially-varying lighting estimation in the outdoor scenes\cite{wang2022neural,SOLD-Net} was introduced. Because they focus on outdoor street scenes, they cannot clearly reproduce the indirect lighting by the scene material. 

\noindent\textbf{Multi-view inverse rendering and neural rendering.}
Earlier works such as Intrinsic3D~\cite{maier2017intrinsic3d} and Zhang~\etal \cite{zhang2016emptying} perform multi-view inverse rendering without deep learning, but require additional equipment to obtain RGB-D images. 3D geometry-based methods~\cite{cis-texture, invpath, kim2016multi} require additional computation to generate mesh. PhotoScene~\cite{photoscene}, in particular, requires external CAD geometry, which should be manually aligned for each object. Moreover, all of these methods require test-time optimization. In contrast, we only need per-view depth maps and we does not require 3D geometry or test-time optimization, making it more computationally efficient and much easier to apply to more general scenes. Philip \etal\cite{philip2019, philip2021} demonstrated a successful relighting with multi-view images; however, these methods cannot be used for applications such as object insertion because they use trained neural renderers. The advent of NeRF\cite{nerf, mipnerf, mip360} has led to a breakthrough in the field of neural rendering research. Several recent methods\cite{nerd, nerv, physg, zhang2021nerfactor, refnerf, invrender2022} have successfully performed inverse rendering using multi-view images; however, they are difficult to apply to scene-level inverse rendering because they are only trained and tested on object-centric images.

\section{Method}
\label{sec:method}
In this section, we describe the detailed architecture of the proposed network, MAIR. Let $K$ denote the number of viewpoints; then, the inputs to the network are $K$ triples, where each triple is composed of an RGB image with $H$,$W$ size ($\mathrm{I} \in \mathbb{R}^{3 \times H \times W}$), depth map ($\Tilde{\mathrm{D}} \in \mathbb{R}^{H \times W}$), and its confidence map ($\Tilde{\mathrm{C}} \in \mathbb{R}^{H \times W}$). $\Tilde{\mathrm{D}}$ and $\Tilde{\mathrm{C}}$ are obtained using a state-of-the-art MVS model\cite{giang2021curvature}. We designed a three-stage structure that progressively estimates the normal, direct lighting, material, and spatially-varying lighting. The entire MAIR pipeline is summarized in Fig.~\ref{fig:whole}.

\subsection{Stage 1 - Target View Analysis Stage}
Stage 1 of MAIR comprises three estimation networks: Normal map (NormalNet), \textbf{In}cident \textbf{D}irect \textbf{L}ighting (InDLNet), and \textbf{Ex}itant \textbf{D}irect \textbf{L}ighting  (ExDLNet). Inspired by recent studies \cite{cis2020, vsg, lighthouse}, we adopt spatially-varying spherical Gaussians (SVSGs)~\cite{cis2020} and volumetric spherical Gaussian (VSG)~\cite{vsg} for the representation of incident lighting and exitant lighting, respectively. 

\noindent{\bf Normal map estimation.} 
Unlike single-view-based methods\cite{cis2020, irisformer2022, Li22, vsg}, where normal information should be inferred from the scene context, the normal map ($\Tilde{\mathrm{N}}$) can be directly derived from the depth map. Thus, NormalNet can show robust performance especially for real-world images, where the distribution of image contents and geometry largely differs from the training data. Still, use of other available information, including the RGB, depth gradient map ($\nabla\Tilde{\mathrm{D}} \in \mathbb{R}^{H \times W}$), and confidence map, can help NormalNet better handle unreliable depth predictions. Specifically, NormalNet is formulated as follows:
\vspace{-2mm} 
\begin{equation}\label{eqn:eq_normal}
\Tilde{\mathrm{N}} = \text{NormalNet}(\mathrm{I},\Tilde{\mathrm{D}}, \nabla\Tilde{\mathrm{D}}, \Tilde{\mathrm{C}}), \Tilde{\mathrm{N}} \in \mathbb{R}^{3 \times H \times W}.
\end{equation}

\begin{figure*}[ht]
  \centering
  \includegraphics[width=\linewidth]{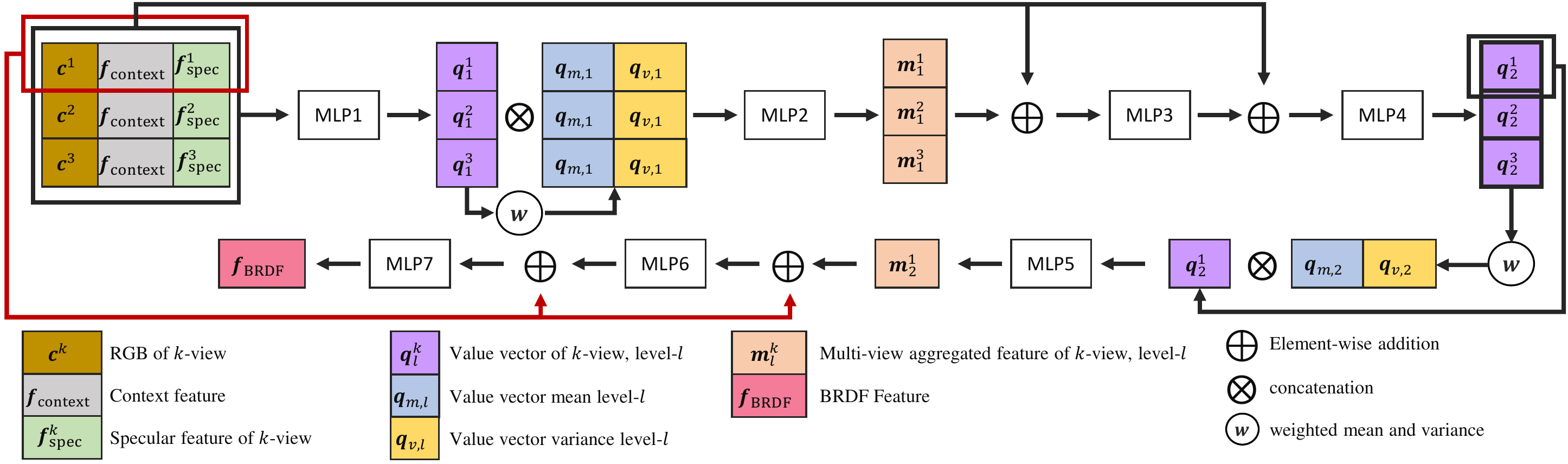}
   \caption{An illustration of MVANet when $K$=3. MVANet creates a value vector by encoding color, context feature, and specular feature, and uses multi-view weights as attention to create multi-view aggregated features. Since our goal is to obtain the BRDF of target-view($1$-view), in level-2, only the value vector of target view is processed.}
   \label{fig:mvanet}
\end{figure*}

\noindent{\bf Incident direct lighting estimation.} 
Given $\mathrm{I}$, $\Tilde{\mathrm{D}}$, $\Tilde{\mathrm{C}}$, and $\Tilde{\mathrm{N}}$ obtained using NormalNet, we estimate SVSGs as a lighting representation of incident direct lighting, which is proven to be effective in modeling environment map~\cite{cis2020}. The proposed InDLNet is formulated as follows: 

\vspace{-2mm} 
\begin{equation}
\{\boldsymbol{\xi}_{s}\}, \{\lambda_{s} \}, \{\boldsymbol{\eta}_{s} \} = \text{InDLNet}(\mathrm{I}, \Tilde{\mathrm{N}}, \Tilde{\mathrm{D}}, \Tilde{\mathrm{C}}),
\label{eqn:incident_SGs}
\end{equation}

where $\boldsymbol{\xi}_s \in \mathbb{R}^2 $ is the direction vector outward from the center of the unit sphere, $\lambda_s \in \mathbb{R}$ is sharpness, and $\boldsymbol{\eta}_s \in \mathbb{R}^3$ is intensity. The environment map is then parameterized with $S_D$ SG lobes ${\{\boldsymbol{\xi}_s, \lambda_s, \boldsymbol{\eta}_s\}_{s=1}^{S_D}}$.  For the $s$-th SG, its radiance $\mathcal{G}(\boldsymbol{l})$ in the direction $\boldsymbol{l} \in \mathbb{R}^2$ can be obtained as

\vspace{-2mm} 
\begin{equation}
\mathcal{G}(\boldsymbol{l};\boldsymbol{\eta}_s,\lambda_s,\boldsymbol{\xi}_s)=\boldsymbol{\eta}e^{\lambda_s(\boldsymbol{l}\cdot \boldsymbol{\xi}_s -1)},
\end{equation}
Using all $S_D$ SG lobes, the incident radiance $\mathcal{R}_{i}(\boldsymbol{l})$ in the direction $\boldsymbol{l}$  is expressed as

\vspace{-2mm} 
\begin{equation}
\mathcal{R}_{i}(\boldsymbol{l}) = \displaystyle\sum_{s=1}^{S_D} \mathcal{G}(\boldsymbol{l};\boldsymbol{\eta}_{s},\lambda_{s},\boldsymbol{\xi}_{s}),
\end{equation}
Li \etal\cite{cis2020} used $S_D=12$ to represent complex spatially-varying lighting; however, we found $S_D=3$ to be sufficient to model much simpler direct lighting. Also we used global intensity to make the SVSGs spatially coherent.

\noindent{\bf Exitant direct lighting estimation.} 
Although effective, the above environment map alone is insufficient to model lighting in a 3D space. Thus, we adopt a voxel-based representation called VSG~\cite{vsg} to further model exitant direct lighting. ExDLNet estimates exitant direct lighting volume $\Tilde{\mathrm{V}}_\text{DL}$ as

\vspace{-4mm} 
\begin{equation}
\Tilde{\mathrm{V}}_\text{DL} = \text{ExDLNet}(\mathrm{I}, \Tilde{\mathrm{N}}, \Tilde{\mathrm{D}}, \Tilde{\mathrm{C}}), \Tilde{\mathrm{V}}_\text{DL} \in \mathbb{R}^{8 \times X \times Y \times Z},
\end{equation}
where $X$, $Y$, and $Z$ are the sizes of the volume. Each voxel in $\Tilde{\mathrm{V}}_\text{DL}$ contains opacity $\alpha$ and SG parameters ($\boldsymbol{\eta}, \boldsymbol{\xi}, \lambda$). From VSG, alpha compositing in the direction $\boldsymbol{l}$  allows us to calculate the incident radiance $\mathcal{R}_{e}(\boldsymbol{l})$ as follows:
\vspace{-2mm} 
\begin{equation}
\mathcal{R}_{e}(\boldsymbol{l}) = \displaystyle\sum_{n=1}^{N_R}\prod_{m=1}^{n-1} (1-\alpha_m)\alpha_n \mathcal{G}(-\boldsymbol{l};\boldsymbol{\eta}_{n},\lambda_{n},\boldsymbol{\xi}_{n}),
\end{equation}
where $N_R$ is the number of ray samples, and $\boldsymbol{\eta}_{n}$,$\lambda_{n}$, and $\boldsymbol{\xi}_{n}$ are the SG parameters of the sample. 

\begin{figure}[ht]
  \centering
  \includegraphics[width=\linewidth]{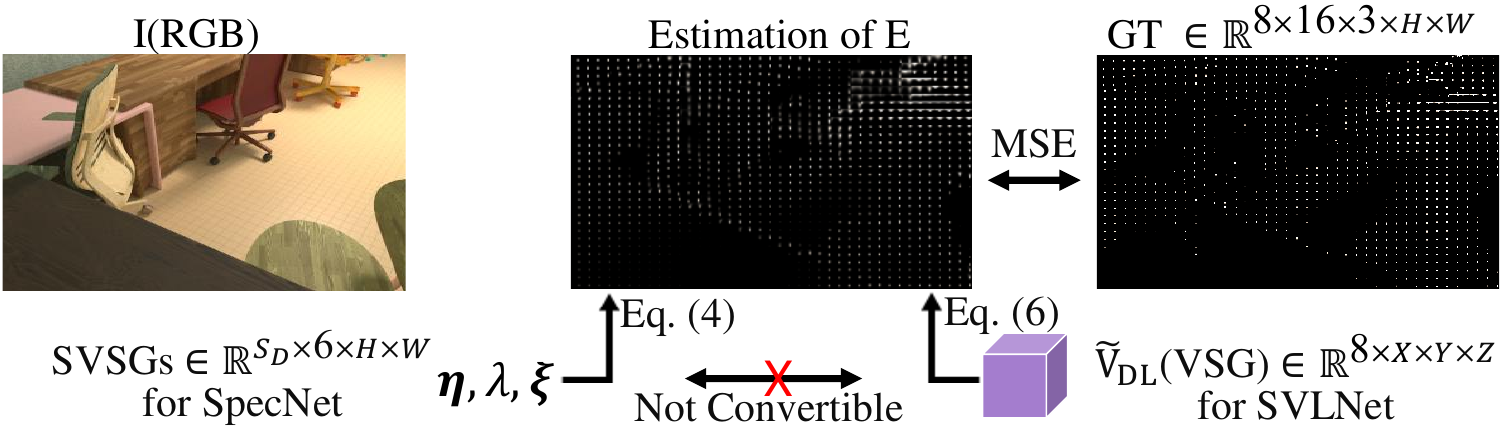}
  \caption{Explanation of the incident/exitant direct lighting. E means per-pixel direct lighting environment map.}
  \label{fig:DL_explain}
\end{figure}

The intensity at which light converges at a point, \ie, incident radiance, serves as guidance for the network to infer diffuse and specular reflections. Meanwhile, information on how light is present in 3D space, \ie, exitant radiance, helps the network infer indirect lighting. Fig.~\ref{fig:DL_explain} is an explanation of the incident/exitant direct lighting. Note that these two representations are not convertible and have different physical-meanings. Please see supplementary for a detailed description of direct lighting.

\vspace{-2mm} 
\vspace{-2mm} 

\subsection{Stage 2 - Material Estimation Stage}
For BRDF estimation, the specular radiance must be considered. To obtain a specular radiance feature $\boldsymbol{f}_{\text{spec}}$, we propose a SpecNet. The diffuse radiance $\boldsymbol{\kappa}$ and specular radiance $\boldsymbol{\gamma}$ of the microfacet BRDF model~\cite{microfacet} are as follows:

\vspace{-3mm} 
\begin{equation}
\boldsymbol{\kappa} = \frac{\boldsymbol{a}}{\pi}\int_{\boldsymbol{l}} L(\boldsymbol{l})\boldsymbol{n}\cdot\boldsymbol{l} \,d\boldsymbol{l}, \boldsymbol{\kappa} \in \mathbb{R}^3
\label{eqn:diffuse_brdf}
\end{equation}

\vspace{-2mm} 
\begin{equation}
\boldsymbol{\gamma} = \int_{\boldsymbol{l}} L(\boldsymbol{l})\mathcal{B}_s(\boldsymbol{v}, \boldsymbol{l}, \boldsymbol{n}, r) \boldsymbol{n}\cdot\boldsymbol{l} \,d\boldsymbol{l}, \boldsymbol{\gamma} \in \mathbb{R}^3
\label{eqn:specular_brdf}
\end{equation}

where $\boldsymbol{a}$ is diffuse albedo, $\boldsymbol{l}$ is lighting direction, $L(\boldsymbol{l})$ is lighting intensity, $\mathcal{B}_s$ is specular BRDF, $\boldsymbol{n}$ is normal, $r$ is roughness, and $\boldsymbol{v}$ is viewing direction, respectively. Since \cref{eqn:specular_brdf} is highly complicated, it is necessary to efficiently encode inputs to make it easier for the network to learn. To this intent, we first rewrite the arguments of $\mathcal{B}_s$ as following:

\vspace{-2mm} 
\begin{equation}
[\mathcal{F}(\boldsymbol{v},\boldsymbol{h}), (\boldsymbol{n}\cdot \boldsymbol{h})^2, \boldsymbol{n}\cdot \boldsymbol{l}, \boldsymbol{n}\cdot \boldsymbol{v}, r],
\end{equation}

where $\mathcal{F}$ is the Fresnel equation, and $\boldsymbol{h}$ is the half vector. Then, we approximate the lighting of \cref{eqn:specular_brdf} with the SVSGs of \cref{eqn:incident_SGs}. 
Since each SG lobe ${\{\boldsymbol{\xi}, \lambda, \boldsymbol{\eta} \}}$ can be thought of as an individual light source, $\boldsymbol{\xi}$, $\boldsymbol{\eta}$, and $\lambda$ can be regarded as $\boldsymbol{l}$, $L(\boldsymbol{l})$, and a parameter to approximate the integral, respectively. Consequently, $\boldsymbol{\gamma}$ can be read as follows:

\vspace{-3mm} 
\small
\begin{equation}
\boldsymbol{\gamma} = {\sum\limits_{s=1}^{S_D} g(\mathcal{F}(\boldsymbol{v},\boldsymbol{h}_s), (\boldsymbol{n}\cdot \boldsymbol{h}_s)^2, \boldsymbol{n}\cdot \boldsymbol{\xi}_s, \boldsymbol{n}\cdot \boldsymbol{v}, \boldsymbol{\eta}_s, \lambda_s, r)},
\label{eqn:specular_approx}
\end{equation}
\normalsize
where $g$ is a newly defined function from our reparameterization. Using \cref{eqn:specular_approx}, we define SpecNet as follows:

\vspace{-5mm} 
\begin{multline}\label{eqn:fspec}
\boldsymbol{f}_{\text{spec}}^{k} = \displaystyle\sum_{s=1}^{S_D} m_s\text{SpecNet}(\mathcal{F}(\boldsymbol{v}_k,\boldsymbol{h}_{s,k}), (\Tilde{\boldsymbol{n}}\cdot \boldsymbol{h}_{s,k})^2, \\
\Tilde{\boldsymbol{n}} \cdot \boldsymbol{\xi}_s, \Tilde{\boldsymbol{n}} \cdot \boldsymbol{v}_k, \boldsymbol{\eta}_s, \lambda_s), 
\end{multline}
\vspace{-4mm} 
\begin{equation}\label{eqn:eq_mask}
m_s = \begin{cases}
   1 &\text{if } \lVert \boldsymbol{\eta}_s \rVert_1\Tilde{\boldsymbol{n}} \cdot \boldsymbol{\xi}_s > 0, \\
   0 &\text{else, } 
\end{cases}
\end{equation}
where $k$ is $k$-th view. A binary indicator $m_s$ is used to exclude SG lobes from $\boldsymbol{\gamma}$ if the intensity of the light source ($\lVert \boldsymbol{\eta}_s \rVert_1$) is 0 or the dot product of the normal and light axis ($\Tilde{\boldsymbol{n}} \cdot \boldsymbol{\xi}_s$) is less than 0. Since SpecNet approximates \cref{eqn:specular_brdf} with this physically-motivated encoding, $\boldsymbol{f}_{\text{spec}}$ can include feature for specular radiance information.
In addition to SpecNet, we use ContextNet to obtain a context feature map $\boldsymbol{f}_{\text{context}} = \text{ContextNet}(\mathrm{I},\Tilde{\mathrm{D}},\Tilde{\mathrm{C}},\Tilde{\mathrm{N}})$ that contains the local context of the scene. All views share $\boldsymbol{f}_{\text{context}}$ of the target view. 

Next, a \textbf{M}ulti-\textbf{V}iew \textbf{A}ggregation network (MVANet) is used to aggregate $\boldsymbol{f}_\text{spec}$, $\boldsymbol{f}_\text{context}$, and RGB across the pixels from all $K$ views, which corresponds to the target view pixel considering MVS depths. However, some of these pixel values might have negative effect if they are from the wrong surfaces due to occlusion. 
To consider occlusion, the depth projection error in $k$-view, denoted as $e_k = \max(-\log(\mid \Tilde{d}_k - z_k \mid), 0)$, is calculated. $\Tilde{d}_k$ is the depth at the pixel position obtained by projecting a point seen from the target view onto $k$-view, and $z_k$ is the distance between the point and the camera center of $k$-view. The depth projection error $\boldsymbol{e} \in \mathbb{R}^K$ is obtained by aggregating $e_k$ from all $K$ views. We use multi-view weight $\boldsymbol{w} = \frac {\boldsymbol{e}} {||\boldsymbol{e}||_1}, \boldsymbol{w} \in \mathbb{R}^K$ as attention weights during the multi-view feature aggregation in MVANet. Our intuition for material estimation is to consider the mean and variance of RGB. MVANet first encodes the input for each view to produce a value vector $\boldsymbol{q}$, and produces a mean and variance of $\boldsymbol{q}$ according to $\boldsymbol{w}$\cite{ibrnet}. It is encoded again and produces a multi-view aggregated feature $\boldsymbol{m}$. Since $\boldsymbol{m}$ is created from weighted means and variances, it has multi-view information considering occlusion. This process is repeated once again for the target view. See Fig.~\ref{fig:mvanet} for detailed structure of MVANet. 

Since MVANet exploits only local features, long-range interactions within the image need to be further considered for inverse rendering\cite{irisformer2022}. Thus, we propose RefineNet for albedo ($\Tilde{\mathrm{A}} \in \mathbb{R}^{3 \times H \times W}$), roughness($\Tilde{\mathrm{R}} \in \mathbb{R}^{H \times W}$) estimation using $\boldsymbol{f}_{\text{BRDF}}$ from MVANet.

\vspace{-5mm} 
\begin{equation}\label{eqn:eq_refine}
\Tilde{\mathrm{A}}, \Tilde{\mathrm{R}} = \text{RefineNet}(\mathrm{I},\Tilde{\mathrm{D}},\Tilde{\mathrm{C}},\Tilde{\mathrm{N}}, \boldsymbol{f}_{\text{BRDF}}, \boldsymbol{f}_{\text{context}}).
\end{equation}

\subsection{Stage 3 - Lighting Estimation Stage}
In stage 3, \textbf{S}patially \textbf{V}arying \textbf{L}ighting Estimation Network(SVLNet) infers 3D lighting with direct lighting, geometry, and material. To this end, we create a visible surface volume ($\mathrm{T} \in \mathbb{R}^{10 \times X \times Y \times Z}$). Although Wang \etal\cite{vsg} used a similar representation, they used a Lambertian reflectance model, which cannot represent complex lighting due to specularity. In contrast, we initialize $\mathrm{T}$ by reprojecting $\mathrm{I}, \Tilde{\mathrm{N}}, \Tilde{\mathrm{A}}, \Tilde{\mathrm{R}}$, which can model specularity. For each voxel, let $(u, v)$ and $d$ denote the projected coordinate of the center point and the depth, respectively. Then, the local feature $\boldsymbol{t} \in \mathbb{R}^{10}$ for each voxel is initialized as follows.
\begin{equation}
\boldsymbol{t} =\left [\rho \mathrm{I}(u,v), ~\rho \Tilde{\mathrm{N}}(u,v), ~\rho \Tilde{\mathrm{A}}(u,v), ~\rho \Tilde{\mathrm{R}}(u,v)\right ], 
\end{equation}
where $\rho = e^{-\Tilde{\mathrm{C}}(u,v)(d-\Tilde{\mathrm{D}}(u,v))^2}$. Note that the confidence ($\Tilde{\mathrm{C}}$) is used to reflect the accuracy of the depth. $\mathrm{T}$ and $\Tilde{\mathrm{V}}_\text{DL}$ are fed to SVLNet, producing outputs $\Tilde{\mathrm{V}}_\text{SVL}$ representing 3D spatially-varying lighting volume, as follows:
\vspace{-1mm} 
\begin{equation}
\Tilde{\mathrm{V}}_\text{SVL} = \text{SVLNet}(\Tilde{\mathrm{V}}_\text{DL}, \mathrm{T}), \Tilde{\mathrm{V}}_\text{SVL} \in \mathbb{R}^{8 \times X \times Y \times Z}.
\end{equation}
In previous work\cite{vsg}, they used implicit global feature volume, which is unclear what information it contains, but we specified $\Tilde{\mathrm{V}}_\text{DL}$ explicitly so that the network can learn the interaction of light source, material, and geometry. 

\begin{figure*}[ht]
  \centering
  \includegraphics[width=\linewidth]{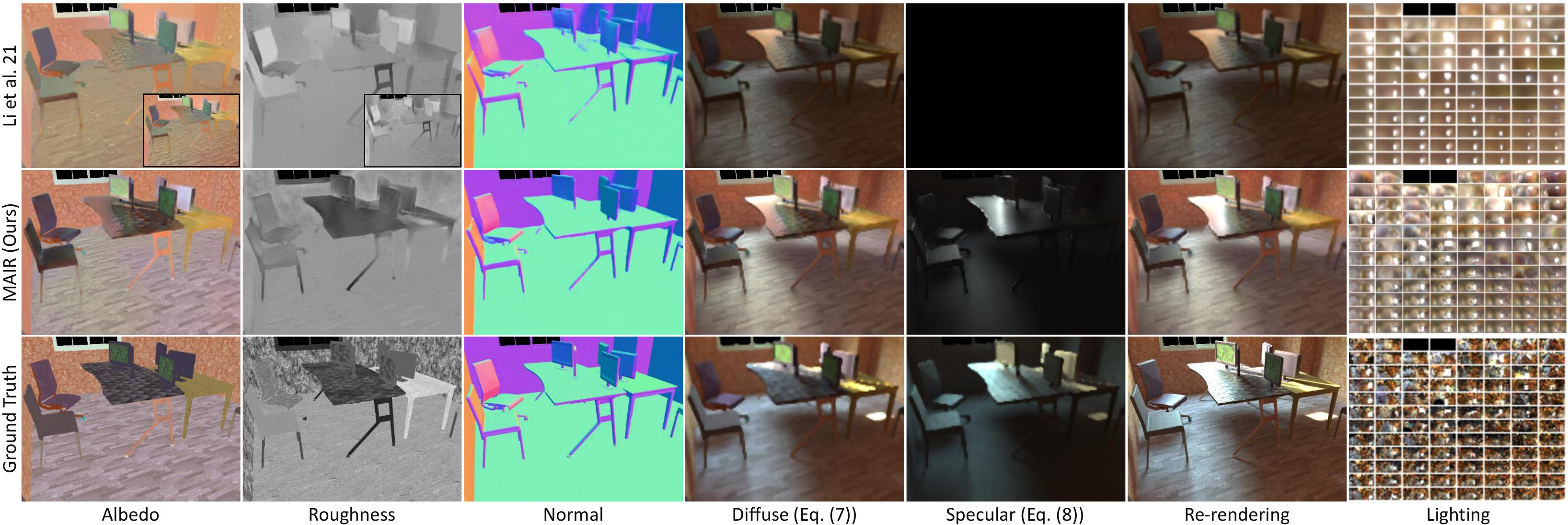}
   \caption{Material, geometry, lighting estimation on OpenRooms FF test data. The small insets in the first row are the material estimation processed by Bilateral Solvers (BS). Even for scenes where inverse rendering is difficult due to strong specular radiance, our method can obtain material, geometry, and lighting more accurately. Please see the lighting and material of the desk.}
   \label{fig:ir_result_synthetic}
\end{figure*}

\begin{figure*}[ht]
  \centering
  \includegraphics[width=\linewidth]{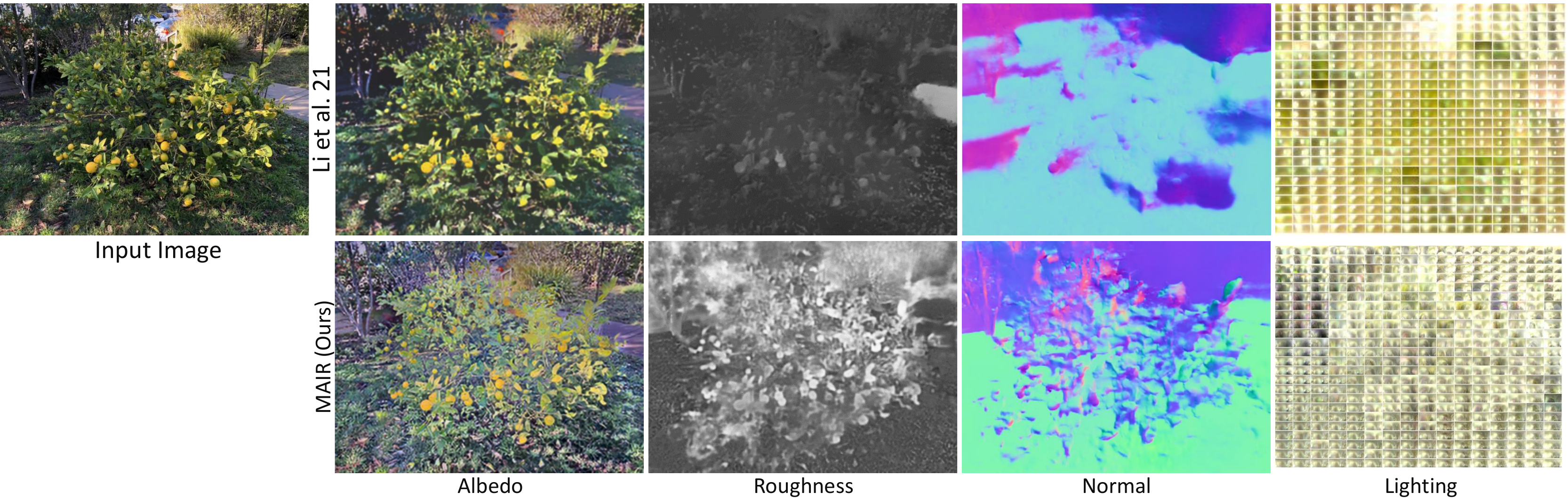}
   \caption{Inverse rendering results for unseen real-world data. Li \etal\cite{cis2020} failed the geometry estimation and predicted the material almost exactly as the input image, while we can remove shadow and disentangle the material and lighting from the image.}
   \label{fig:ir_result_real}
\end{figure*}

\section{Implementation Details}
\label{sec:implementation}
\noindent{\bf Dataset.} 
OpenRooms\cite{openrooms2021} provides many HDR images and various ground truths for indoor scenes; however, the distribution of camera poses is too random to be used in multi-view applications. Therefore, we created OpenRooms FF. Among OpenRooms\cite{openrooms2021} images, 23,618 images were selected to be appropriate for multi-view rendering based on the camera position and minimum depth, and eight neighboring frames were rendered by translating the camera for each image. For a detailed description of OpenRooms FF, please see supplementary material.

\begin{figure*}[t]
  \centering
  \includegraphics[width=\linewidth]{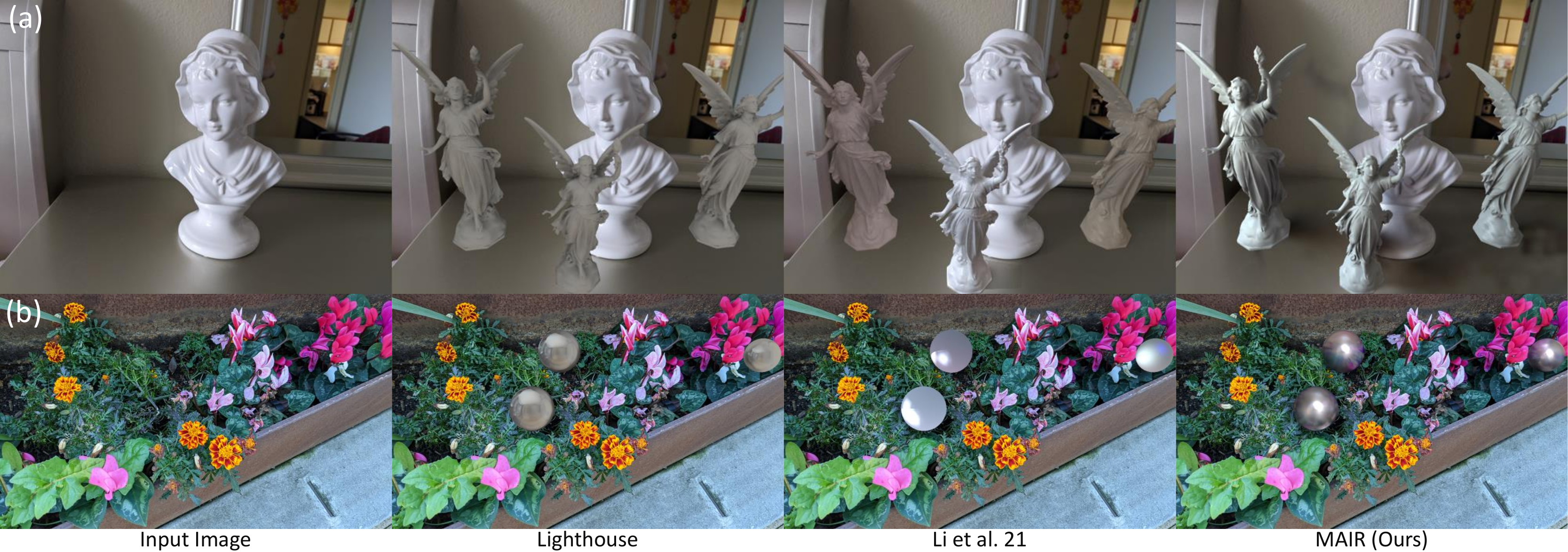}
   \caption{Object~\cite{stanford} and floating chrome sphere insertion comparison for unseen real-world data from IBRNet dataset\cite{ibrnet}. Our method estimate lighting robustly for unseen images, and we can insert objects more realistically than previous methods.}
   \label{fig:oi_real}
\end{figure*}

\noindent{\bf Training and loss.} 
We used nine images ($K$=9) in our experiments. We trained each stage separately since the ground truth required for each stage is available in OpenRooms FF. We trained lighting with only 2D per-pixel ground truth lighting, but our VSG can represent lighting well in any 3D space (See (b) of Fig.~\ref{fig:oi_real}.). A brief description of the loss for each stage is provided in Tab.~\ref{tab:loss}. For a detailed description of training, loss function, and network architecture, please see supplementary material.

\begin{table}[ht]
\footnotesize
\centering
\begin{tabular}{|c|l|l|}
\hline
\multicolumn{1}{|c|}{Stage} & \multicolumn{1}{c|}{Network} & \multicolumn{1}{c|}{Loss}\\
\hline
\multirow{3}{*}{1} & NormalNet & MSE($\mathrm{N}$) + L1 angular($\mathrm{N})$\\
& InDLNet & si-MSE(Lighting)  \\ 
& ExDLNet & si-MSE(Lighting) \\ 
\hline
2 & All & si-MSE($\mathrm{A}$) + MSE($\mathrm{R}$) \\
\hline
3 & All & si-MSE(Lighting) + MSE(Re-rendering)\\
\hline
\end{tabular}
\caption{Training loss for MAIR. si- means scale invariant.}
\vspace{-3mm}
\label{tab:loss}
\end{table}

\section{Experiments}
\label{sec:experiments}
Since no previous works, to the best of our best knowledge, addressed multi-view scene-level inverse rendering, we compare our method qualitatively and quantitatively with a single-view-based method\cite{cis2020}. We also conduct a qualitative performance evaluation on the real-world dataset~\cite{ibrnet}. Last, we demonstrate that our realistic 3D spatially-varying lighting through virtual object insertion.

\begin{figure*}[t]
  \centering
  \includegraphics[width=\linewidth]{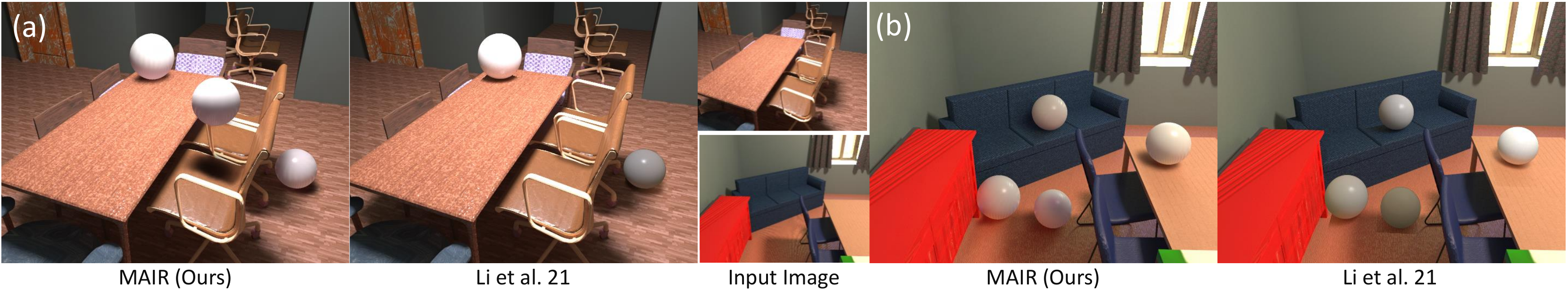}
   \caption{Object insertion comparison for OpenRooms FF test scene. Our method can insert the sphere to match the scene shadow, reproduce color bleeding, and put the sphere in the air.}
   \label{fig:oi_result_synthetic}
\end{figure*}

\subsection{Evaluation on Synthetic Data}
For quantitative comparison, we trained Li \etal\cite{cis2020} on our OpenRooms FF. Tab.~\ref{tab:compare} shows the performance comparison for all test images in OpenRooms FF. Material and geometry errors were measured by MSE and lighting error was measured by log-space MSE with ground truth per-pixel environment map. Our method outperformed the previous method in geometry and material estimation. Moreover, we achieved better lighting performance even though the previous method directly estimates per-pixel lighting but we obtain per-pixel lighting from the 3d lighting volume. For the re-rendering, the previous method showed better performance on average because Li \etal\cite{cis2020}'s cascade structure self-supervises re-rendering through the rendering layer. However, in scenes where specular radiance appears strongly, we found that our method robustly and accurately estimates the details of lighting and materials, thereby yielding more realistic result, as shown in Fig.~\ref{fig:ir_result_synthetic}. 

\begin{table}[ht]
\newcommand{\B}{\fontseries{b}\selectfont} 
\footnotesize
\centering
\begin{tabular}{|l|S[detect-weight]|S[detect-weight]S[explicit-sign=+]|} 
\hline
MSE ($\times 10^{-2}$)& \multicolumn{1}{c|}{Li \etal \cite{cis2020}} & \multicolumn{2}{c|}{MAIR (Ours)}\\
\hline
Albedo $\downarrow$ & 0.569 & \B 0.368 &{(}-0.201{)}\\
Normal $\downarrow$ & 2.71 & \B 1.36 &{(}-1.35{)}\\
Roughness $\downarrow$ & 3.66 & \B 2.70 &{(}-0.96{)}\\
Lighting $\downarrow$ & 13.74 & \B 12.04 &{(}-1.70{)}\\
Re-rendering $\downarrow$ & \B 0.554 & 0.633 &{(}+0.079{)}\\
\hline
\end{tabular}
\caption{Quantitative comparison of material, geometry, and lighting in OpenRooms FF. Albedo and roughness of Li \etal\cite{cis2020} are processed with a Bilateral Solver (BS).}
\label{tab:compare}
\vspace{-3mm}
\end{table}

\subsection{Evaluation on Real-world Data}
We evaluate the performance of inverse rendering for the real-world scenes in the IBRNet dataset\cite{ibrnet} in which the scene context is difficult to grasp. As shown in Fig.~\ref{fig:ir_result_real}, the single-view-based method fails in geometry estimation due to the lack of the scene context, leading to the inaccurate disentanglement of the material and illumination. Our method, on the other hand, obtains a reasonable geometry, removes shadows, and predicts spatially-consistent albedo. More results can be found in the supplementary material.

\subsection{Evaluation on Object Insertion}
We further demonstrate the effectiveness of our robust inverse rendering on the object insertion task. In Fig.~\ref{fig:oi_result_synthetic},~\ref{fig:oi_real} we provide a comparison with the single-view method~\cite{cis2020} and the stereo method\cite{lighthouse}. It should be noted that, we use the normal from MAIR for Lighthouse\cite{lighthouse} results because Lighthouse\cite{lighthouse} does not provide scene geometry. In Li \etal\cite{cis2020}'s object insertion application, the user must specify the plane on which the object is located. Therefore, object cannot be placed on another object or floated in the air, and shadows are only cast on the plane. In our method, it is possible to insert objects freely into the 3D space without restrictions on the shadow. We also conducted a user study on the object insertion, in which respondents chose the most realistic images rendered by competing methods for 25 scenes. As shown in the Tab.~\ref{tab:oi_userstudy}, our method shows the best performance in both synthetic and real-world scenes.

\begin{table}[ht]
\footnotesize
\centering
\resizebox{\linewidth}{!}{
\begin{tabular}{|l|c|c|c|}
\hline
Scene & \multicolumn{1}{|c|}{Lighthouse~\cite{lighthouse}} & \multicolumn{1}{|c|}{Li \etal\cite{cis2020}} & \multicolumn{1}{|c|}{MAIR}\\
\hline
OpenRooms FF & - &0.220  &\textbf{0.780}\\
IBRNet~\cite{ibrnet}& 0.120  &0.180  & \textbf{0.700}\\
IBRNet~\cite{ibrnet}(chrome)  & 0.186  & 0.025 & \textbf{0.789} \\
\hline
\end{tabular}
}
\caption{User study results on virtual object insertion. The design and details are described in supplementary material.}
\label{tab:oi_userstudy}
\vspace{-2mm}
\end{table}

\noindent{\bf Object insertion in indoor test dataset.} In (a) of Fig.~\ref{fig:oi_result_synthetic}, the sphere on the floor is inserted to match the geometry and lighting of the scene, and the sphere in the air is inserted to express the appropriate shadows for the scene geometry. In (b) of Fig.~\ref{fig:oi_result_synthetic}, the sphere inserted on the floor represents the color bleeding, showing that our method is capable of indirect lighting estimation.

\noindent{\bf Object insertion in real-world unseen dataset.} To show the generalization ability of the proposed model trained only on the synthetic indoor images, we tested object insertion on the real-world unseen dataset\cite{ibrnet}. In (a) of Fig.~\ref{fig:oi_real}, Li \etal\cite{cis2020} failed to separate lighting from the material, and as a result, the appearance of the object was colored with the material on the floor. Our method successfully separate material and lighting from the image, and realistic lighting is represented on the inserted object. Lighthouse\cite{lighthouse} was more vulnerable to unseen real-world data because it did not consider scene geometry and materials, even though they use stereo images. Their objects have a similar appearance regardless of the scene. In (b) Fig.~\ref{fig:oi_real}, we inserted a chrome sphere to validate the indirect lighting estimation accuracy of the proposed method. Because lighthouse\cite{lighthouse} was trained with an indoor LDR dataset, it always tended to create an environment map with an indoor scene on the sphere, and failed to reproduce HDR lighting. Li \etal\cite{cis2020}'s pixel lighting was not sufficient to express indirect lighting of the scene reflected in chrome sphere. In contrast, our chrome sphere shows realistic lighting that reflects the surrounding environment. Please refer to the supplementary material for more results.

\subsection{Ablation Study}
\noindent{\bf Design of stage 2.} 
Tab.~\ref{tab:abl_DL_stage2} shows experimental results for network design choices in stage 2. It shows that MVANet requires a local context of ContextNet for material estimation. In addition, RefineNet is necessary to compensate for the weakness of the pixel-wise operation in MVANet. We also validate the effect of $\boldsymbol{f}_{\text{spec}}$.``w/o $\boldsymbol{f}_{\text{spec}}$'' is the result of the model trained without $\boldsymbol{f}_{\text{spec}}$, and ``w/o reparameterize'' is the result of the model trained without physically-motivated encoding as follows:

\vspace{-2mm} 
\begin{equation}\label{eqn:fspec_abl}
\boldsymbol{f}_{\text{spec}}^{k} = \displaystyle\sum_{s=1}^{S_D} \text{SpecNet}(\boldsymbol{v}_k, \Tilde{\boldsymbol{n}}, \boldsymbol{\xi}_s, \boldsymbol{\eta}_s, \lambda_s).
\end{equation}
Improvement in roughness estimation implies that our physically-motivated encoding considering the microfacet BRDF~\cite{microfacet} model helps in estimating specular radiance. 

\vspace{-2mm}
\begin{table}[ht]
\footnotesize
\centering
\begin{tabular}{|l|c|c|}
\hline
MSE ($\times 10^{-2}$)& \multicolumn{1}{|c|}{Albedo$\downarrow$} & \multicolumn{1}{|c|}{Roughness$\downarrow$} \\
\hline
w/o ContextNet &0.498 &4.064  \\
w/o RefineNet &0.661  &3.971  \\
w/o $\boldsymbol{f}_{\text{spec}}$ & 0.436  & 3.056 \\
w/o reparameterize & 0.428 & 3.085 \\
Ours & \textbf{0.423} & \textbf{2.739} \\
\hline
\end{tabular}
\caption{ablation studies in stage 2.}
\label{tab:abl_DL_stage2}
\vspace{-4mm}
\end{table}

\begin{figure}[ht]
  \centering
  \includegraphics[width=\linewidth]{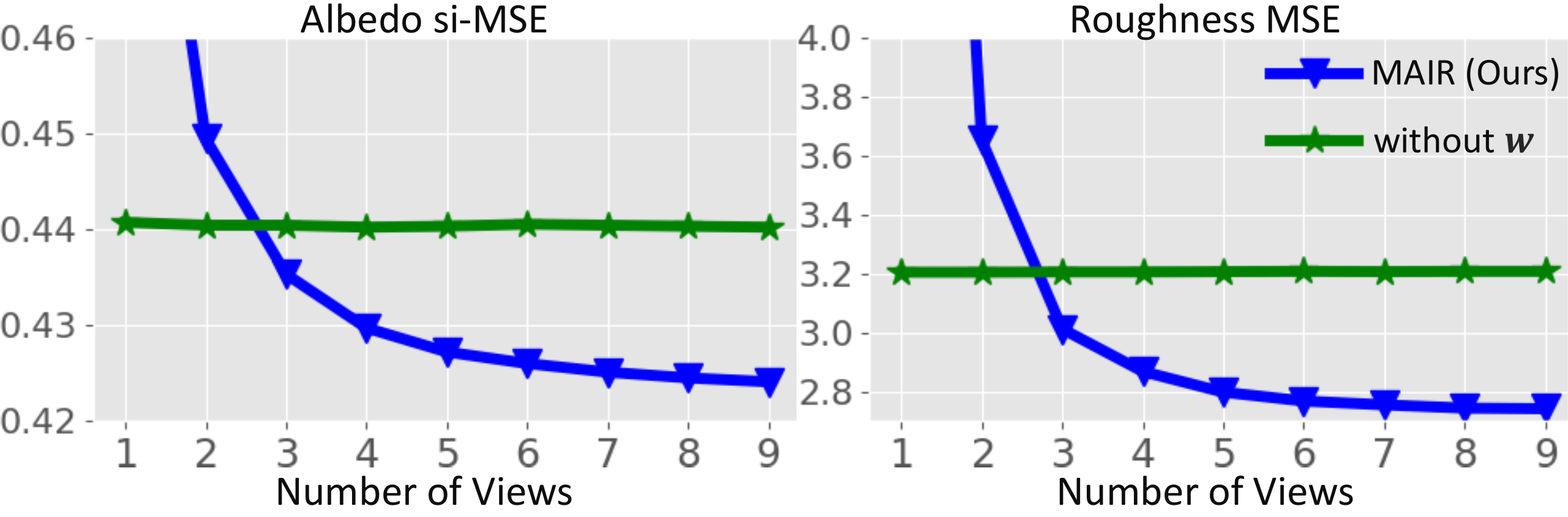}
  \caption{material evaluation depending on the number of views.}
  \label{fig:graph_numview}
\end{figure}

\noindent{\bf Attention with multi-view weight.} We also validate the effect of multi-view weight($\boldsymbol{w}$) in MVANet. Fig.~\ref{fig:graph_numview} shows the material estimation accuracy according to the number of views. When training without $\boldsymbol{w}$, increasing the number of views seems ineffective, possibly due to the negative effect of noises in multi-view features introduced by occlusion. On the other hand, MVANet can selectively focus on the information required for material estimation with $\boldsymbol{w}$, making the accuracy increase with the number of views.

\vspace{-2mm}
\vspace{-2mm}

\begin{figure}[ht]
  \centering
  \includegraphics[width=\linewidth]{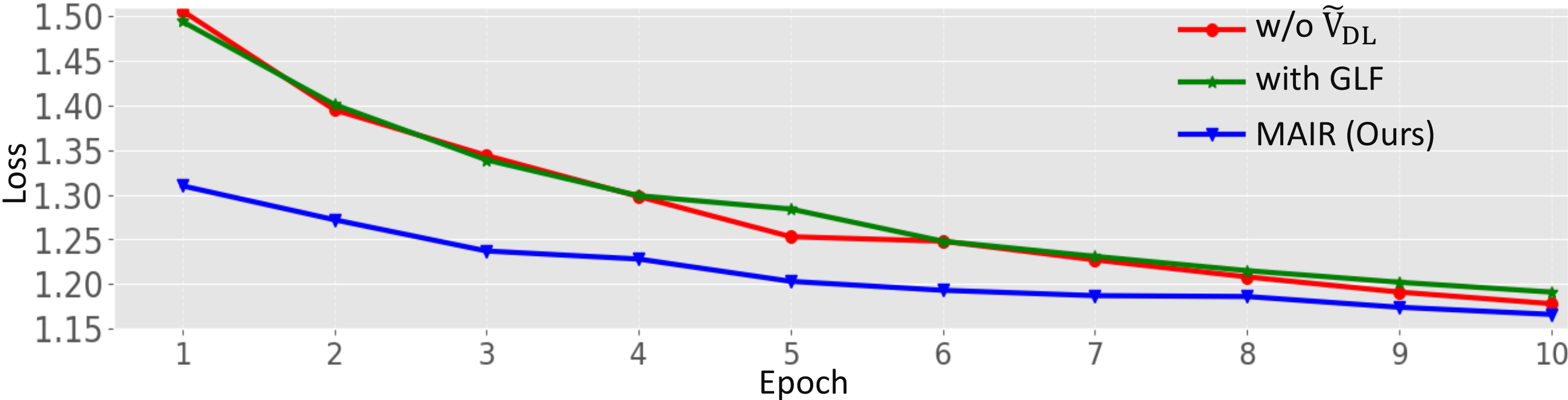}
  \caption{lighting evaluation depending on the input source.}
  \label{fig:graph_vdl}
\end{figure}

\noindent{\bf The necessity of exitant direct lighting.}
Fig.~\ref{fig:graph_vdl} shows that $\Tilde{\text{V}}_\text{DL}$ is helpful for indirect lighting estimation. The result of training with a global lighting feature as in \cite{vsg}, instead of $\Tilde{\text{V}}_\text{DL}$, is also compared (with GLF). The $\Tilde{\text{V}}_\text{DL}$ helps network to converge quickly, which indicates that $\Tilde{\text{V}}_\text{DL}$ helps the network infers indirect lighting.

\vspace{-2mm}

\section{Discussion}
\begin{figure}[ht]
  \centering
  \includegraphics[width=\linewidth]{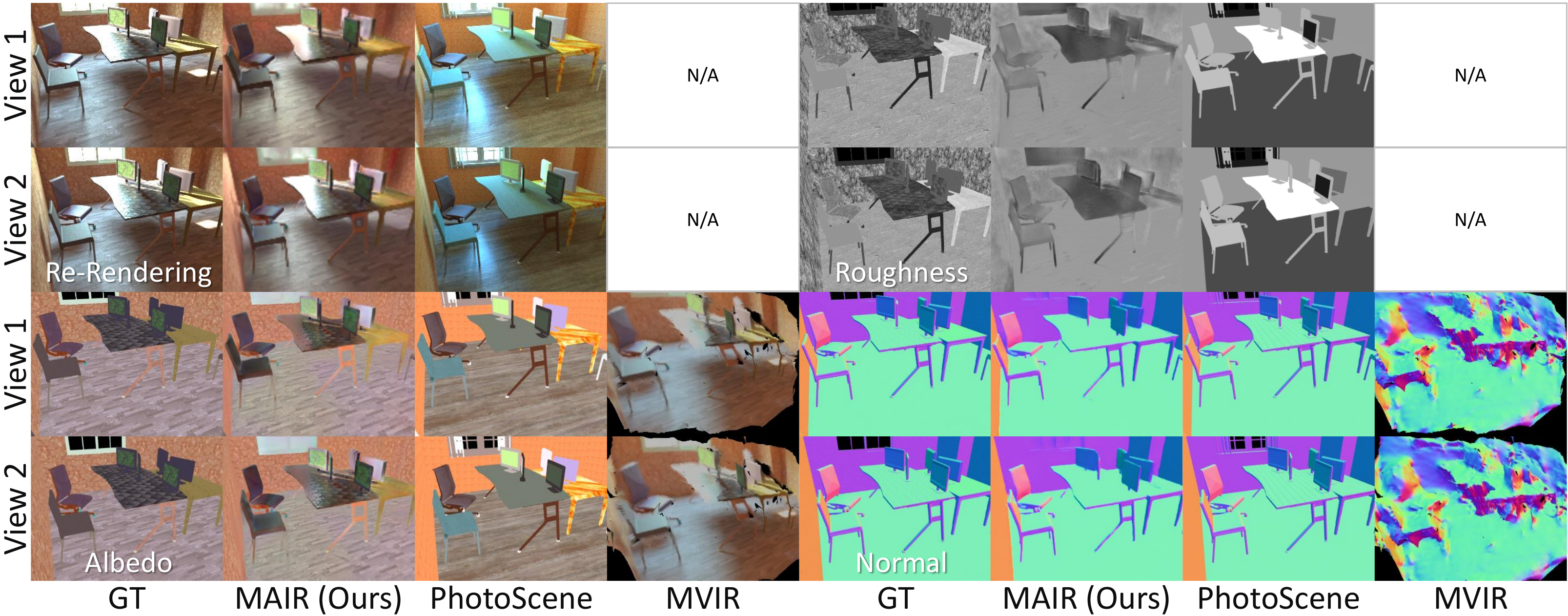}
  \caption{Comparisons with PhotoScene~\cite{photoscene} and MVIR~\cite{kim2016multi}.}
  \label{fig:comparisons}
\end{figure}

\noindent{\bf Comparisons with 3D geometry-based methods.}
Fig.~\ref{fig:comparisons} shows the comparison results with PhotoScene~\cite{photoscene} and MVIR~\cite{kim2016multi} in 2 views. The inference time is 2s for ours, 9m 52s for MVIR~\cite{kim2016multi} and 10m 40s for PhotoScene~\cite{photoscene} on RTX 2080 Ti. MVIR~\cite{kim2016multi} fails to generate geometry, showing severe artifacts. The PhotoScene~\cite{photoscene} shows a complete scene reconstruction thanks to the CAD geometry and material graph, but the synthesized images largely differ from the original scene. 

\noindent{\bf Inter-view consistency.}
Unlike 3D geometry-based methods~\cite{photoscene, kim2016multi}, which leverage global 3D geometry to achieve consistency, Our MAIR operates in pixel-space and therefore does not explicitly guarantee inter-view consistency. Still, we did not observe significant inconsistencies during our experiment, presumably due to the role of MVANet, which can be seen in Fig.~\ref{fig:comparisons}. For object insertion, we used lighting of the center-view.

\vspace{-2mm}
\vspace{-2mm}

\begin{figure}[ht]
  \centering
  \includegraphics[width=\linewidth]{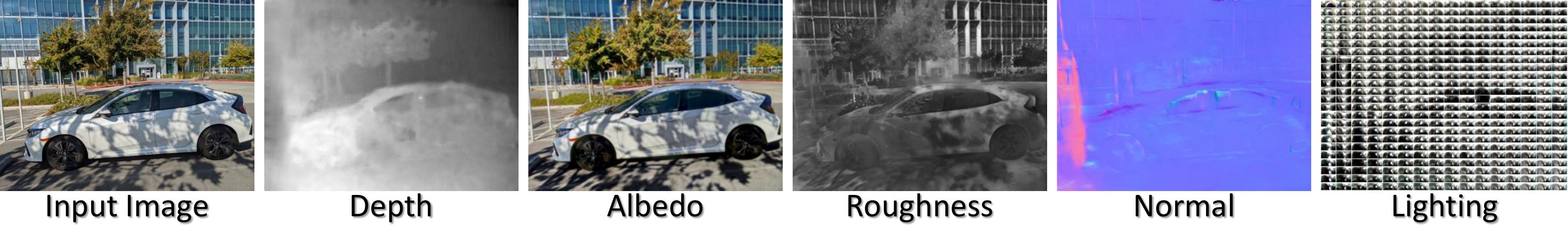}
  \caption{Failure case when depth prediction fails.}
  \label{fig:failure_case}
\end{figure}

\noindent{\bf Limitation.}
One possible limitation comes from the cascaded nature of the pipeline. If depth estimation fails due to, for example, presence of dynamic objects or large textureless region, our MAIR will not work properly (See Fig.~\ref{fig:failure_case}.). Another possible limitation comes from the VSG representation. Although VSG can express 3D lighting effectively, it cannot be applied to applications such as light source editing because it is non-parametric. 

\noindent{\bf Conclusion.}
We presented the first practical multi-view scene-level inverse rendering method by creating an multi-view HDR synthetic dataset. Compared to the single-view based methods, we found that our method is more robust for unseen real-world scenes, providing high-quality virtual object insertion results. We believe that our work can elevate image-based rendering and physically-based rendering together, so realizing a higher level of inverse rendering and scene reconstruction.

\textbf{{\huge Appendix}}

\appendix

\section{Appendix Outline}
These appendices provide details about the OpenRooms FF dataset (Appendix~\ref{B}), details about direct lighting (Appendix~\ref{C}) and analysis of lighting estimation results (Appendix~\ref{D}), view synthesis applications (Appendix~\ref{E}), additional implementation details (Appendix~\ref{F}), and additional experimental results (Appendix~\ref{G}).

\section{OpenRooms FF dataset}\label{B}
\begin{figure}[ht]
  \centering
  \includegraphics[width=\linewidth]{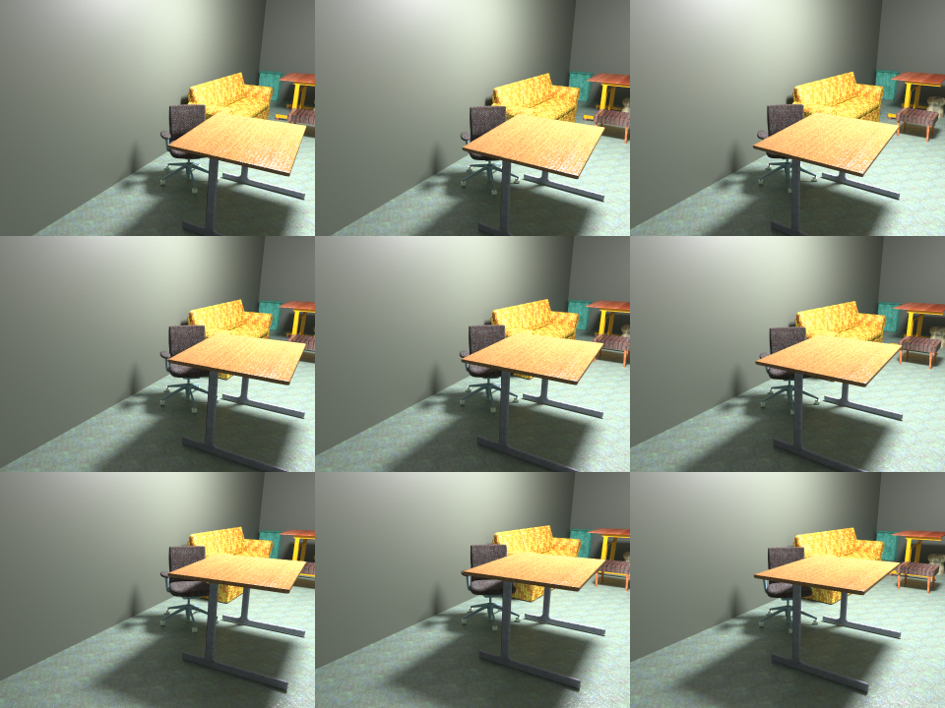}
   \caption{Sample of forward facing multi-view images in OpenRooms FF.}
   \label{fig:33}
\end{figure}

We created a dataset for multi-view inverse rendering called OpenRooms Forward Facing (Openrooms FF) dataset. Openrooms FF is an extension of the existing single-view inverse rendering dataset, OpenRooms~\cite{openrooms2021}, and most of resources to build the dataset are provided by the authors of OpenRooms~\cite{openrooms2021}, including data sources and creation tools. The materials, however, were unavailable due to the licensing issue, so we had to purchase materials from Adobe Stock~\cite{adobestock} except for 200 materials that were not found from Adobe Stock; instead, we replace them with other similar materials. We selected 23,618 images from the OpenRooms dataset by filtering out the images in which the camera looks at a wall or window, lacks textures in the scene, or object is too close to the camera. Then, we rendered forward facing multi-view images of 3 $\times$ 3 arrays by moving camera in eight directions: up, right up, right, right down, down, left down, and left, left up using the OptiX-based renderer~\cite{optixrenderer}. The baseline was set proportionally to the average depth of the scene to observe the change in the specular radiance. See Fig.~\ref{fig:33} for a multi-view images sample. As a result, a total of 212,562 (9 $\times$ 23,618) images were created and 27,000 (9 $\times$ 3000) images were separated into test dataset. OpenRooms FF consists of HDR RGB images, diffuse albedo images, roughness images, normal maps, binary masks, depth maps, per-pixel environment maps. We rendered images at 640 $\times$ 480 resolution but resized to 320 $\times$ 240 with bilinear interpolation for the training/test. The OpenRooms FF is summarized in Tab.~\ref{tab:openroomsFF}.

\begin{table}[ht]
\footnotesize
\centering
\begin{tabular}{|l|c|c|}
\hline
& Dataset & Training / Test \\
\hline
HDR RGB & 640 $\times$ 480  & 320 $\times$ 240 \\
Diffuse Albedo & 640 $\times$ 480  & 320 $\times$ 240 \\
Roughness & 640 $\times$ 480  & 320 $\times$ 240 \\
Normal & 640 $\times$ 480  & 320 $\times$ 240 \\
Mask & 640 $\times$ 480  & 320 $\times$ 240 \\
Depth & 640 $\times$ 480  & Not used \\
per-pixel DL & 40 $\times$ 30 $\times$ 32 $\times$ 16  & 40 $\times$ 30 $\times$ 16 $\times$ 8 \\
per-pixel SVL & 160 $\times$ 120 $\times$ 32 $\times$ 16  & 160 $\times$ 120 $\times$ 16 $\times$ 8 \\
\hline
\end{tabular}
\caption{Data type and resolution of OpenRooms FF. Spatially-varying lighting (SVL) has a spatial resolution of 160 $\times$ 120 and an angular resolution of 32 $\times$ 16.}
\label{tab:openroomsFF}
\end{table}

\section{Direct Lighting Details}\label{C}
\begin{figure*}[ht]
  \centering
  \includegraphics[width=\linewidth]{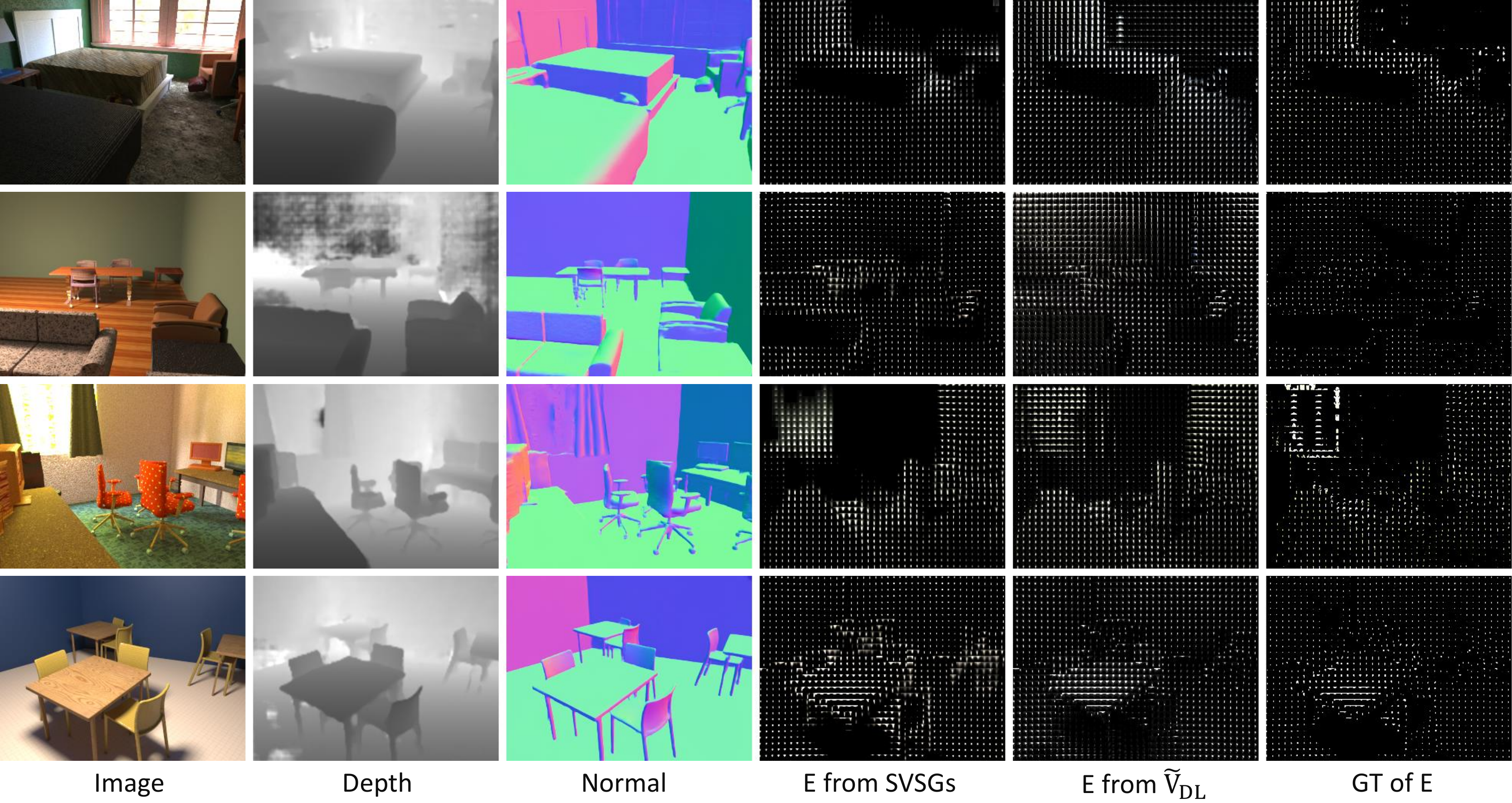}
   \caption{Direct lighting environment map~($16\times8\times3$) estimation results for OpenRooms FF.} 
   \label{fig:DL_example}
\end{figure*}

Since the intensity($\boldsymbol{\eta}_s$) of incident direct lighting is the intensity of the light source, it is unrelated to pixel location. Thus we use global intensities ${\boldsymbol{\eta}_s}$ rather than per-pixel intensities. Instead, per-pixel visibility ${\mu_s \in \mathbb{R}}$ was used to account for occlusion. To enhance the dynamic range of the SG lobes, we use the non-linear transformation~\cite{cis2020}. The ablation study results for $S_D$ in SVSGs of incident direct lighting are shown in Tab.~\ref{tab:abl_SD}. Please see \cref{eqn:eq_lossDL} for $\mathcal{L}_{\text{reg}}$. Direct lighting performance improved as $S_D$ increased, but GPU Memory also increased. We chose $S_D=3$ considering its performance and GPU usage. Fig.~\ref{fig:DL_example} shows the incident(SVSGs) / exitant($\Tilde{\mathrm{V}}_\text{DL}$) direct lighting estimation results. SVSGs generally performed better because $\Tilde{\mathrm{V}}_\text{DL}$ estimates 3D volume, while SVSGs directly estimates 2D per-pixel environment map(E). Also, even though the consistency between them is not considered, since they are trained with the same ground truth(GT), they are consistent enough as shown in the Fig.~\ref{fig:DL_example}.

\begin{table}[ht]
\footnotesize
\centering
\begin{tabular}{|c|c|c|c|}
        \hline
         $S_D$ & si-MSE & {$\mathcal{L}_{\text{reg}}$} & GPU Memory(GB). \\
         \hline
        1 & 0.106 & 0.136 & 10.8\\ 
        \hline
        2 & 0.103 &0.127 & 11.26\\
        \hline
        \textbf{3} & \textbf{0.101} & \textbf{0.092} & \textbf{12.72}\\
        \hline
        4 & 0.101 & 0.081 & 13.43\\
        \hline
        6 & 0.100 & 0.061 & 14.94\\
        \hline
\end{tabular}
\caption{The ablation study results for $S_D$ in SVSGs.}
\label{tab:abl_SD}
\end{table}

\section{Analysis of Lighting Estimation Results}\label{D}
\begin{figure*}[ht]
  \centering
  \includegraphics[width=\linewidth]{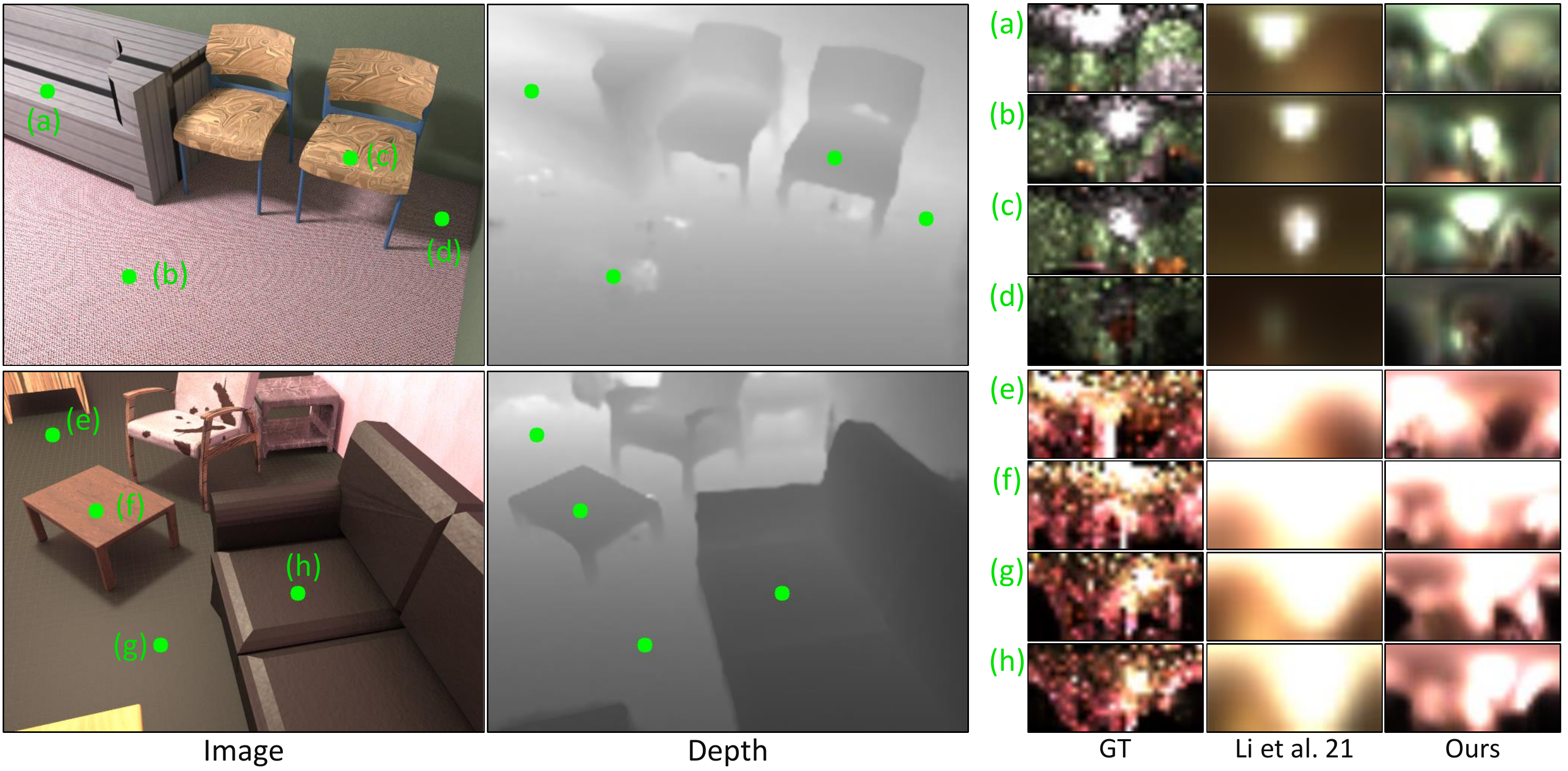}
   \caption{Per-pixel environment map~($32\times16\times3$) estimation results for OpenRooms FF.}
   \label{fig:L_analysis}
\end{figure*}

We have analyzed spatially-varying lighting quality in detail. Since the SVLNet implementation is quite memory-hungry, the resolution of our $\Tilde{\mathrm{V}}_\text{SVL}$ is $128^3$, which is low compared to the image resolution ($320\times240$ ). Also, because the field-of-view of our camera setup is limited, the lighting of the out-of-view area must rely on context inference about the dataset. Fig.~\ref{fig:L_analysis} shows the per-pixel lighting estimation results for the OpenRooms FF test scene. In the Fig.~\ref{fig:L_analysis}, our estimation approximates the overall outline of the GT better than Li~\etal\cite{cis2020} , but fails to mimic the high frequency details of the GT due to limitations in resolution and field-of-view.

\section{View Synthesis}\label{E}
While image-based rendering(IBR) can perform view interpolation excellently, the view-dependent effect of highly specular objects, such as chrome spheres, is difficult to reproduce using IBR. Physically-based rendering(PBR) can handle this view-dependent effect realistically, but PBR requires scene material, geometry, and spatially-varying lighting that is difficult to obtain in the real-world. Because MAIR can perform accurate inverse rendering in real-world scenes, and can be easily applied to existing view synthesis methods with multi-view images, we can take advantage of IBR and PBR. The view synthesis result of the scene with chrome sphere inserted is in the accompanied video. This application consists of two steps: (1) background rendering with NeRF~\cite{nerf}, and (2) object and mask rendering with our renderer. We render the shadow of an object in all images and we train NeRF with these images. Background including shadow in novel view is rendered with NeRF, and chrome sphere in novel view is rendered with our lighting and renderer. Among the variants of NeRF, we use DirectVoxGO~\cite{dvgo} for fast training.

\section{Implementation details}\label{F}
\noindent{\bf Training and architecture details.} Our experiments were conducted with 8 NVIDIA RTX A5000 (24GB). In training, we use Adam optimizer, and the binary mask image $(\mathrm{M}_o, \mathrm{M}_l)$. $\mathrm{M}_o \in \mathbb{R}^{H \times W}$ is mask on pixels of valid materials, and $\mathrm{M}_l \in \mathbb{R}^{H \times W}$ is mask on pixels of valid materials and area lighting. The binary mask image is included in the OpenRooms FF and is used only for training. First, we define masked L1 angular error function ($g_1$), masked MSE function ($g_2$), masked scale invariant MSE function ($g_3$), masked scale invariant $\log$ space MSE function ($g_4$), and regularization function ($g_5$) as follows.

\small
\begin{align}
g_1(A, B, M) = ||(\cos^{-1}(A \odot B)) \otimes M||_1, \\
g_2(A, B, M) = ||{(A - B) \otimes M}||_2^2, \\
g_3(A, B, M) = ||{(A - \tau B) \otimes M}||_2^2, \\
g_4(A, B, M) = ||(\text{log}(A+1) - \text{log}(\tau B+1)) \otimes M) ||_2^2, \\
g_5(A) = -A\log(A),
\end{align}
\normalsize
where $\odot$ is element-wise dot product, $\otimes$ is element-wise multiplication, and $\tau$ is the scale obtained by least square regression between A and B.

In stage 1, the loss function of NormalNet is as follows:
\small
\begin{equation}
\mathcal{L}_{\text{normal}} = \beta_1 g_1(\mathrm{N}, \Tilde{\mathrm{N}}, \mathrm{M}_l)
+ \beta_2 g_2(\mathrm{N}, \Tilde{\mathrm{N}}, \mathrm{M}_l).
\end{equation}
\normalsize
NormalNet has a U-Net\cite{unet} structure with 6 down-up convolution blocks. 

Since the light source is not transparent, we use a regularization $g_5$ so that the visibility $\mu_s$ of InDLNet and the opacity $\alpha$ of ExDLNet can be 0 or 1. the loss function of InDLNet and ExDLNet is as follows:
\small
\begin{align} \label{eqn:eq_lossDL}
\mathcal{L}_{\text{InDL}} = \beta_1 g_4(\mathrm{E}_{DL}, \Tilde{\mathrm{E}}_{DL}, \mathrm{M}_o) +\beta_2 g_5(\mu_s), \\
\mathcal{L}_{\text{ExDL}} = \beta_1 g_4(\mathrm{E}_{DL}, \Tilde{\mathrm{E}}_{DL}, \mathrm{M}_o) +\beta_2 g_5(\alpha),
\end{align} 
\normalsize
where $\mathrm{E}_{DL}$ is the per-pixel direct lighting environment map. InDLNet also has a U-Net structure that encoder is shared, and decoders are separated by $\lambda_s, \xi_s, \mu_s$. The light source intensity $\eta_s$ was decoded using MLP. ExDLNet follows structure of OccNet\cite{occupancy} and uses MLP with conditional batch normalization (CBN) \cite{de2017modulating}. All convolution blocks use batch normalization(BN).

In stage2, the loss function is as follows. 

\small
\begin{equation}
\mathcal{L}_{\text{BRDF}} = \beta_1 g_3(\mathrm{A}, \Tilde{\mathrm{A}}, \mathrm{M}_o) + \beta_2 g_2(\mathrm{R}, \Tilde{\mathrm{R}}, \mathrm{M}_o).
\end{equation}
\normalsize

ContextNet uses U-Net with ResNet18\cite{resnet}, SpecNet uses MLP with 3 layers, MVANet uses layer normalization (LN), and RefineNet uses U-Net with group normalization(GN).

In stage3, the loss function is as follows. 
\small
\begin{multline} 
    \mathcal{L}_{\text{SVL}} = \beta_1 g_4(\mathrm{E}_{SVL}, \Tilde{\mathrm{E}}_{SVL}, \mathrm{M}_o) + \beta_2 g_5(\alpha) \\
    + \beta_3\displaystyle\sum_{k=1}^K ||w_k{(\mathrm{I}^k- \tau_{diff}\Tilde{\mathrm{I}}_{diff} - \tau_{spec}\Tilde{\mathrm{I}}^k_{spec}) \otimes \mathrm{M}_o}||_2^2,
\end{multline} 
\normalsize

where $\mathrm{E}_{SVL}$ is the per-pixel lighting environment map, $\tau_{diff}$ and $\tau_{spec}$ are the scale obtained by least square regression with target image. $\mathrm{I}^k, \Tilde{\mathrm{I}}_{diff}, \Tilde{\mathrm{I}}^k_{spec}$ are $k$-view image, diffuse image, $k$-view specular image, respectively, and $w_k$ is multi-view weight. In SVLNet, visible surface volume ($\mathrm{T}$) is concatenated with $\Tilde{\mathrm{V}}_\text{DL}$ after 2 downsampling and processed with 3D U-Net. The resolution of the $\Tilde{\mathrm{V}}_\text{DL}$ is $32^3$, and the resolution of the $\mathrm{T}$ and $\Tilde{\mathrm{V}}_\text{SVL}$ is $128^3$. SVLNet uses instance normalization(IN). SVLNet needs a lot of memory when training, so we render environment map with a spatial resolution of 60$\times$80. A summary of training, number of GPUs, hyperparameter and network architecture is provided in Tab.~\ref{tab:arch}. Rendering includes the time to obtain a 60$\times$80$\times$8$\times$16 environment map from VSG and the time to re-render the input image.

\begin{table*}[htb!]
\footnotesize
\centering
\resizebox{\linewidth}{!}{
\begin{tabular}{|c|l|c|c|c|c|c|c|c|c|c|c|c|c|}
\hline
Stage &Network &input & Arch & norm &batch &epoch &$\beta_1$ &$\beta_2$ &$\beta_3$ &lr &training / GPUs &inference &output(channels)\\
\hline
\multirow{3}{*}{1}&NormalNet &$\mathrm{I},\Tilde{\mathrm{D}}, \nabla\Tilde{\mathrm{D}}, \Tilde{\mathrm{C}}$,&U-Net  & BN & 96 & 60 & 1.0 & 1.0 & - &2e-3 &7h / 4 & 3ms &$\Tilde{\mathrm{N}}(3)$  \\

&InDLNet &$\mathrm{I}, \Tilde{\mathrm{N}}, \Tilde{\mathrm{D}}, \Tilde{\mathrm{C}}$& U-Net, MLP &BN &384 &80 &1.0 &1e-3 & - &2e-4 &10h / 4 &5ms &$\boldsymbol{\xi}_s, \lambda_s, \mu_s, {\boldsymbol{\eta}_s}(8)$ \\

&ExDLNet &$\mathrm{I}, \Tilde{\mathrm{N}}, \Tilde{\mathrm{D}}, \Tilde{\mathrm{C}}$& U-Net, MLP &BN &96 &80 &1.0 &1e-4 & - & 1e-4 &1d / 8 & 6ms &$\Tilde{\text{V}}_\text{DL}(8)$\\

\hline
\multirow{4}{*}{2} &ContextNet &$\mathrm{I}, \Tilde{\mathrm{N}}, \Tilde{\mathrm{D}}, \Tilde{\mathrm{C}}$ & Res U-Net  & BN &\multirow{4}{*}{64} &\multirow{4}{*}{40} &\multirow{4}{*}{3.0} &\multirow{4}{*}{1.0} &\multirow{4}{*}{-} &\multirow{4}{*}{1e-4} &\multirow{4}{*}{1d 20h / 8} &\multirow{4}{*}{54ms} &$\boldsymbol{f}_\text{context}(32)$\\
&SpecNet &${\boldsymbol{\xi}_s}, {\lambda_s}, {\mu_s}, {\boldsymbol{\eta}_s}, \boldsymbol{v}, \Tilde{\mathrm{N}}$ & MLP & -  &&&&&&&&&$\boldsymbol{f}_\text{spec}(8)$\\
&MVANet &$\mathrm{I}, \boldsymbol{f}_\text{context}, \boldsymbol{f}_\text{spec}, \boldsymbol{w}$ & - &LN &&&&&&&&&$\boldsymbol{f}_\text{BRDF}(16)$\\
&RefineNet &$\mathrm{I}, \Tilde{\mathrm{N}}, \Tilde{\mathrm{D}}, \Tilde{\mathrm{C}}, \boldsymbol{f}_\text{context}, \boldsymbol{f}_\text{BRDF}$ & U-Net &GN &&&&&&&&&$\Tilde{\mathrm{A}}(3), \Tilde{\mathrm{R}}(1)$\\
\hline
3&SVLNet &$\mathrm{I}, \Tilde{\mathrm{N}}, \Tilde{\mathrm{D}}, \Tilde{\mathrm{C}}, \Tilde{\mathrm{A}}, \Tilde{\mathrm{R}}, \Tilde{\text{V}}_\text{DL}$ &3D U-net &IN &8 &10 &10.0 &1e-2 &1.0 &1e-4 &3d 8h / 8 &11ms &$\Tilde{\text{V}}_\text{SVL}(8)$ \\
\hline
- &Rendering & $\Tilde{\mathrm{N}}, \Tilde{\mathrm{D}}, \Tilde{\mathrm{C}}, \Tilde{\mathrm{A}}, \Tilde{\mathrm{R}}, \Tilde{\text{V}}_\text{SVL}$, &-&-&-&-&-&-&-&- &- &834ms &$\Tilde{\mathrm{I}}(3)$\\
\hline

\end{tabular}
}
\caption{The details of the network architecture, and training. Please refer to the main paper for the architecture of MVANet.}
\label{tab:arch}
\end{table*}

\noindent{\bf Test details.} 
Li \etal\cite{cis2020} and we both used an environment map with an angular resolution of 16 $\times$ 8 during training, but we created an environment map with 32 $\times$ 16 during testing because our VSG was not restricted by resolution. In training, all views are rendered for re-rendering loss, but in testing, only the target view was rendered.

\section{Additional Experimental Results}\label{G}
\subsection{Indoor Synthetic Scenes}
\begin{figure*}[t]
  \centering
  \includegraphics[width=0.97\linewidth]{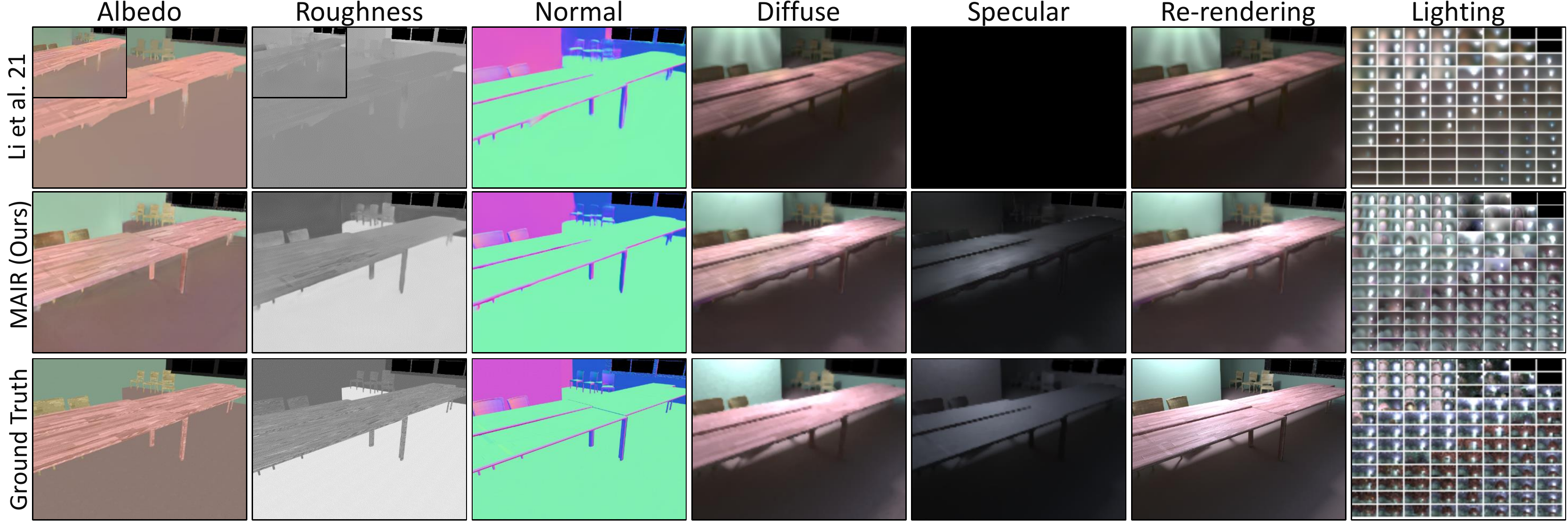}
  \includegraphics[width=0.97\linewidth]{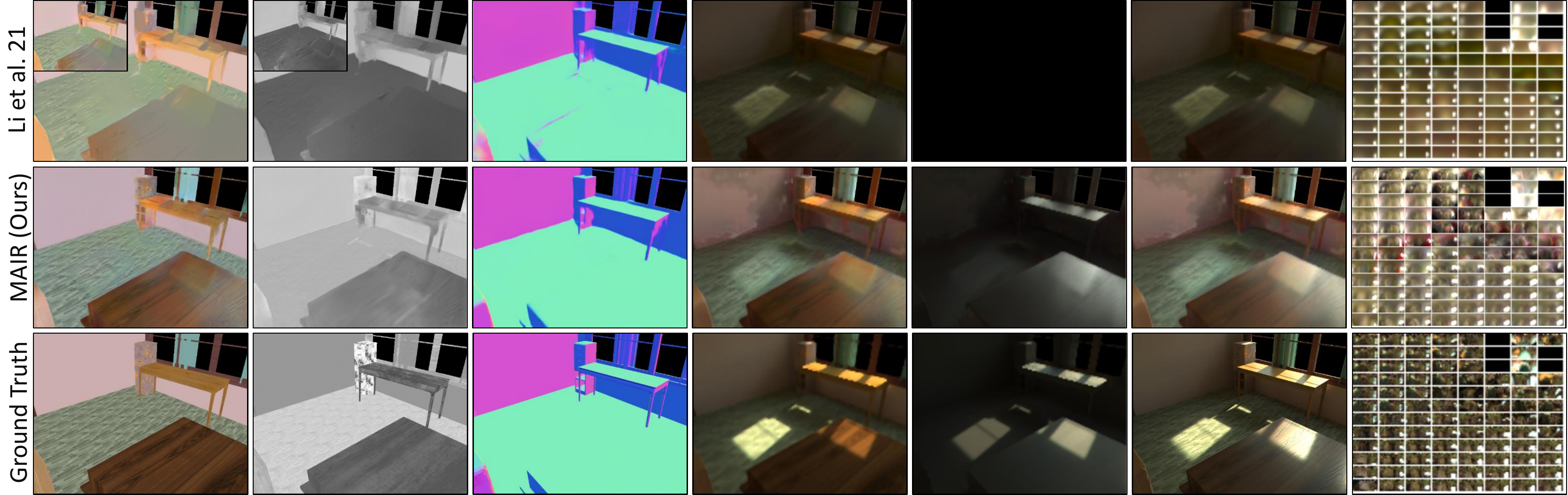}
  \includegraphics[width=0.97\linewidth]{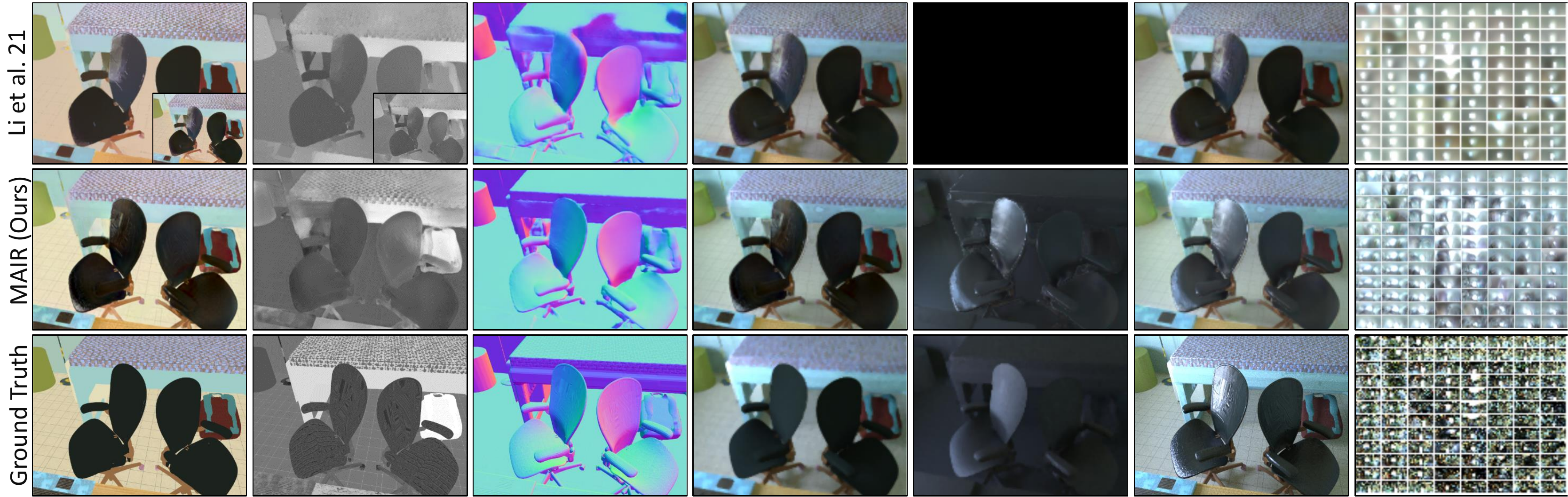}
  \includegraphics[width=0.97\linewidth]{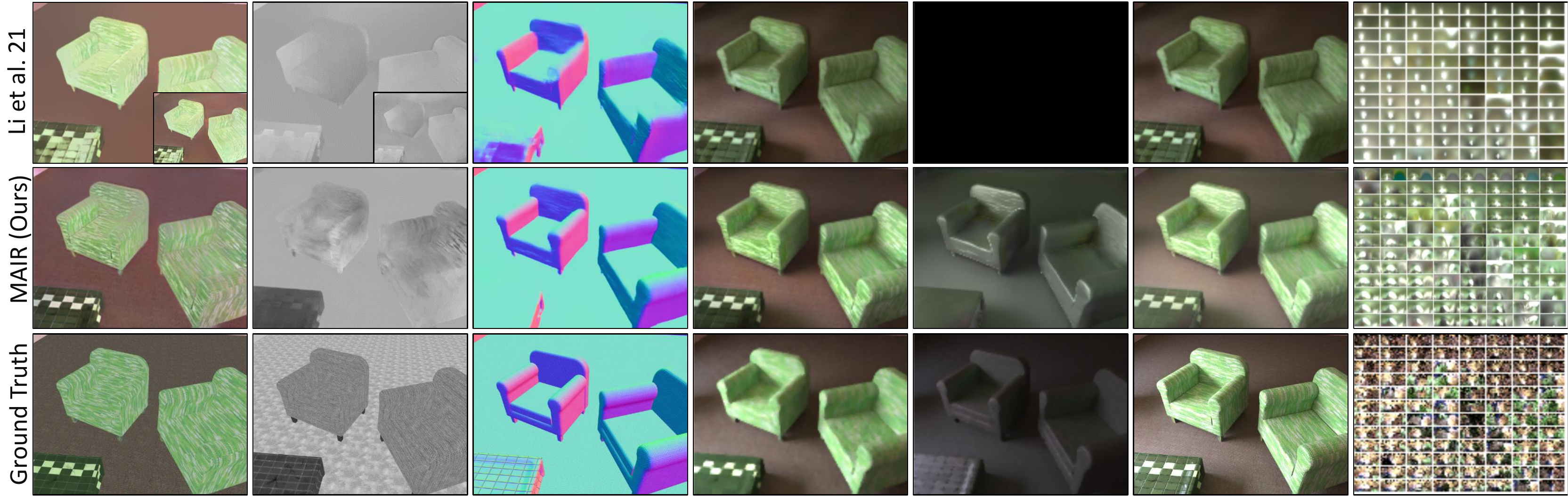}
   \caption{Additional inverse rendering results on OpenRooms FF. Small insets are the estimations without bilateral solver (BS).}
   \label{fig:supp_ir_indoor}
\end{figure*}

We provide additional inverse rendering results for OpenRooms FF test scene in Fig.~\ref{fig:supp_ir_indoor}. Our method leverage multi-view and incident direct lighting to provide more accurate material estimation results for highly specular regions. (\eg table in sample 2, chair in sample 3) Furthermore, the proposed method yields better normal estimation results especially for more complicated structures by utilizing MVS depth. As a result, our lighting is more realistic and we can re-render input image more accurately.

\subsection{Real-World Scenes}
\begin{figure*}[ht]
  \centering
  \includegraphics[width=0.99\linewidth]{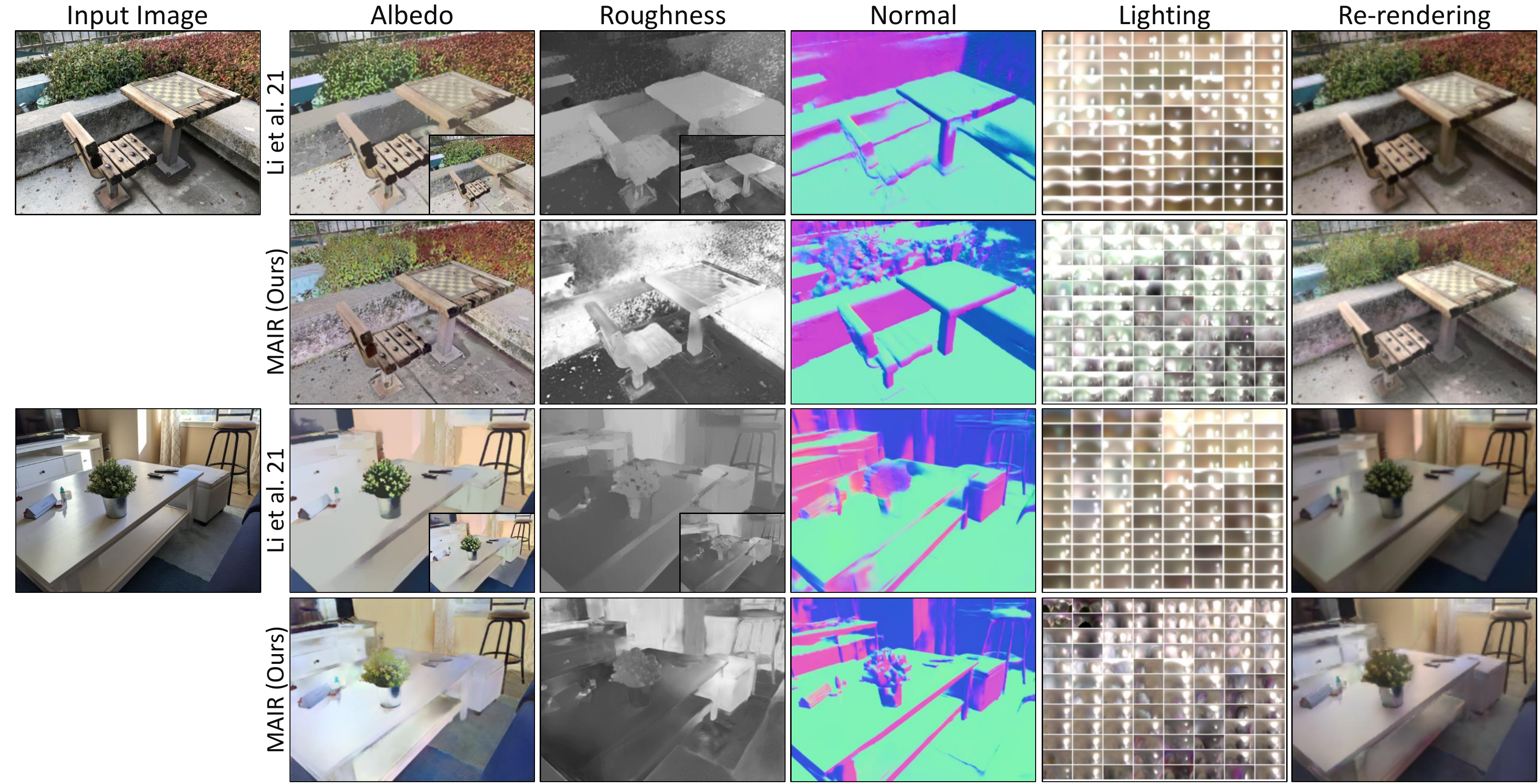}
  \includegraphics[width=0.99\linewidth]{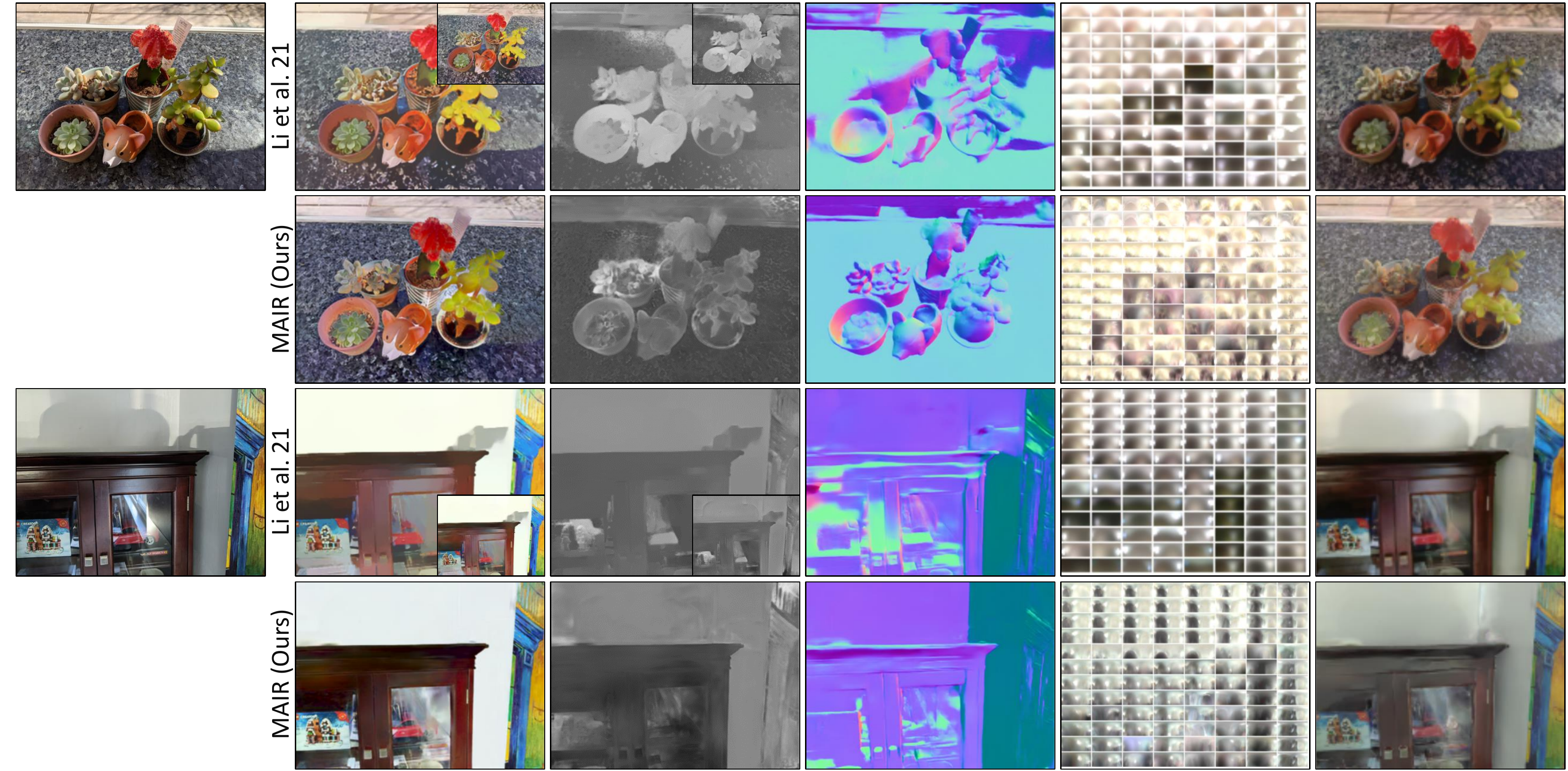}
  \includegraphics[width=0.99\linewidth]{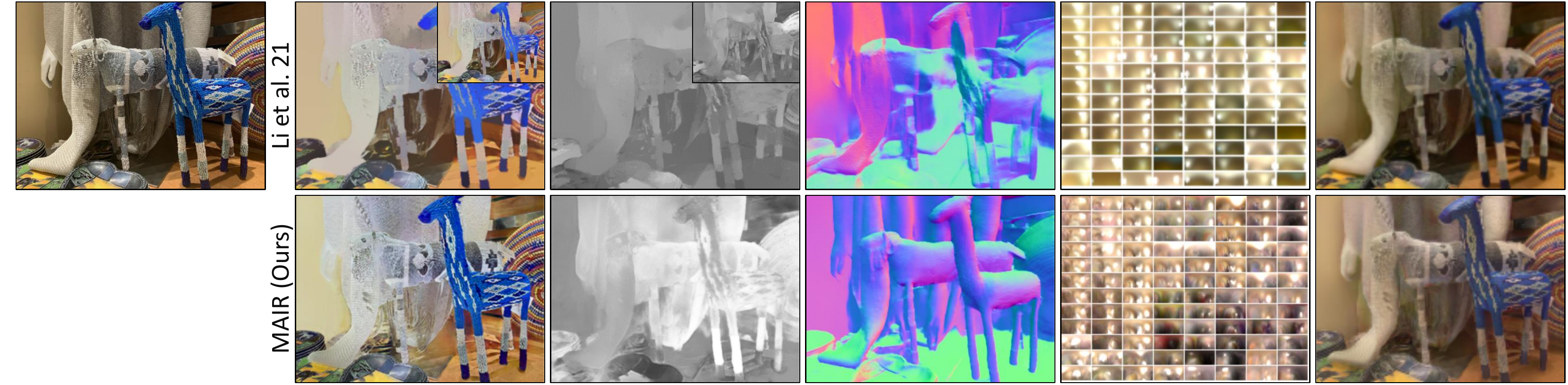}
   \caption{Additional inverse rendering results on IBRNet dataset\cite{ibrnet}. Small insets are the estimations without BS.}
   \label{fig:supp_ir_real}
\end{figure*}

The performance gaps between MAIR and the single-view-based methods are more distinct in the unseen real-world scene. Fig.~\ref{fig:supp_ir_real} shows that our method robustly produces reasonable normal maps even for complex scene structures, and this naturally affects the subsequent material, lighting estimation. MAIR shows better material estimation results for shadowed regions(\eg table, wall in sample 2, floor in sample 3) or specular regions(\eg drawer in sample 4). Although there are no ground truths for materials, from our experience, we know that the stones, bushes in sample 1, and the dolls in sample 5 should show high roughness, which are consistent with our high roughness estimation results.

\subsection{Object Insertion}
Inverse rendering performance of three competing methods, lighthouse~\cite{lighthouse}, Li \etal\cite{cis2020}, and MAIR, are tested by comparing the quality of object insertion. We implemented a simple renderer for object insertion by referring to Wang \etal\cite{vsg} and used it for rendering results of MAIR and lighthouse~\cite{lighthouse}. As the public implementation of Li \etal\cite{cis2020} includes a renderer of their own, results of Li \etal\cite{cis2020} were rendered using this renderer, except for the results of the chrome sphere insertion; the renderer from Li \etal\cite{cis2020} does not support the chrome sphere rendering directly, so we used our renderer for this case. It should be also noted that all results of lighthouse~\cite{lighthouse} were produced by using our scene geometries because scene geometry results from lighthouse~\cite{lighthouse} were not accurate enough to render.

\begin{figure*}[ht]
  \centering
  \includegraphics[width=0.99\linewidth]{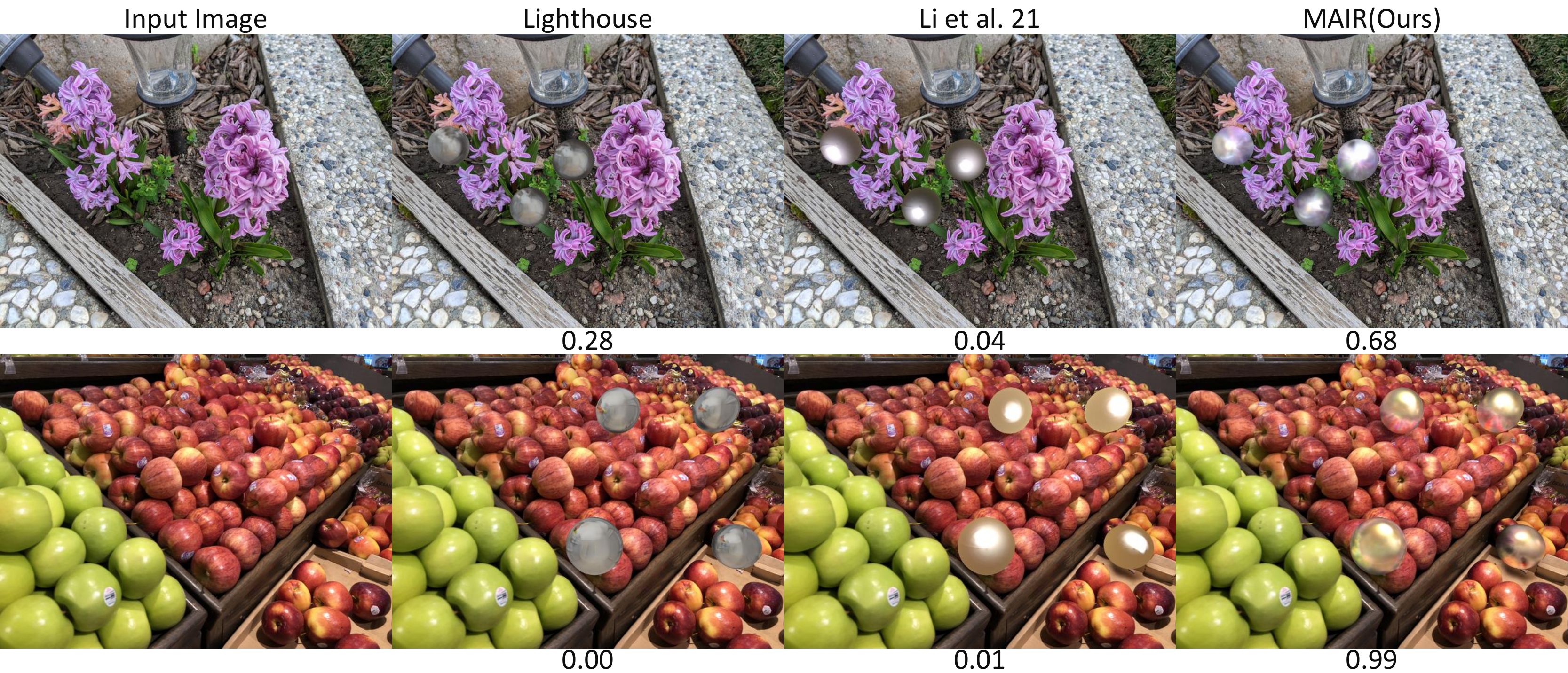}
  \includegraphics[width=0.99\linewidth]{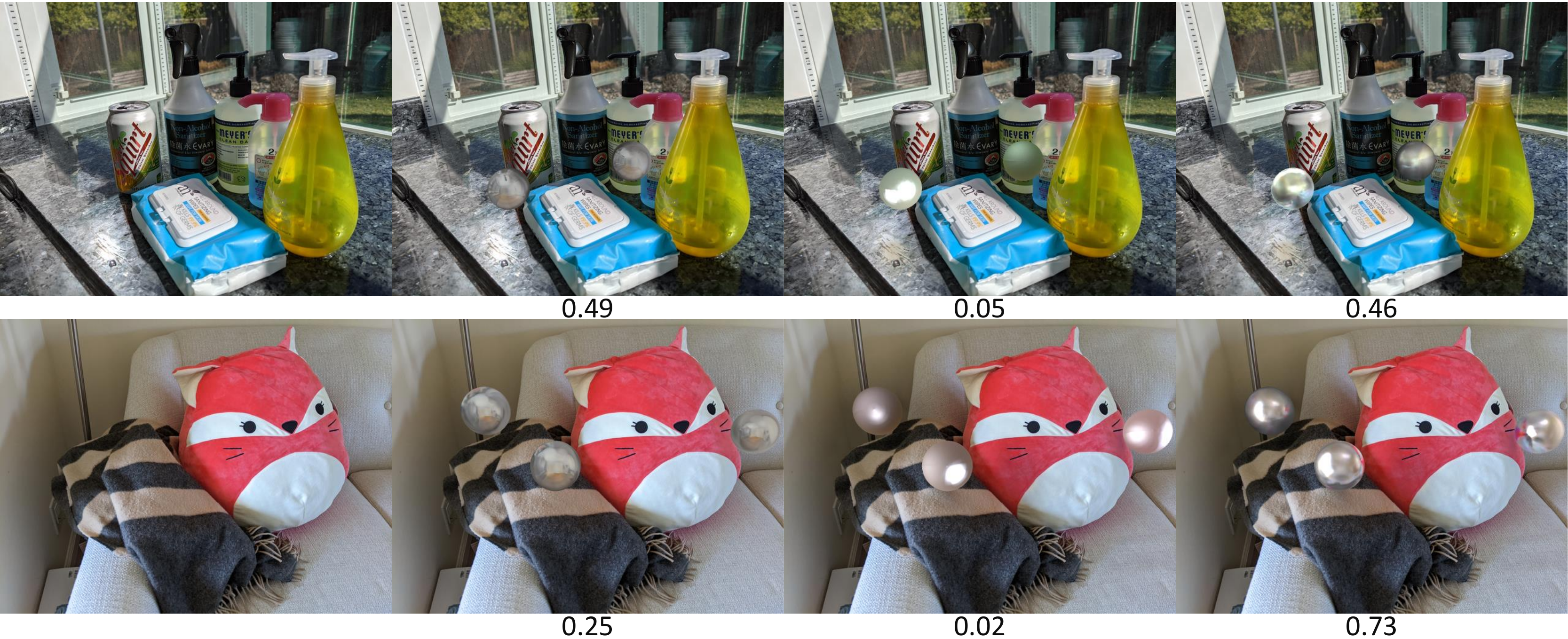}
  \includegraphics[width=0.99\linewidth]{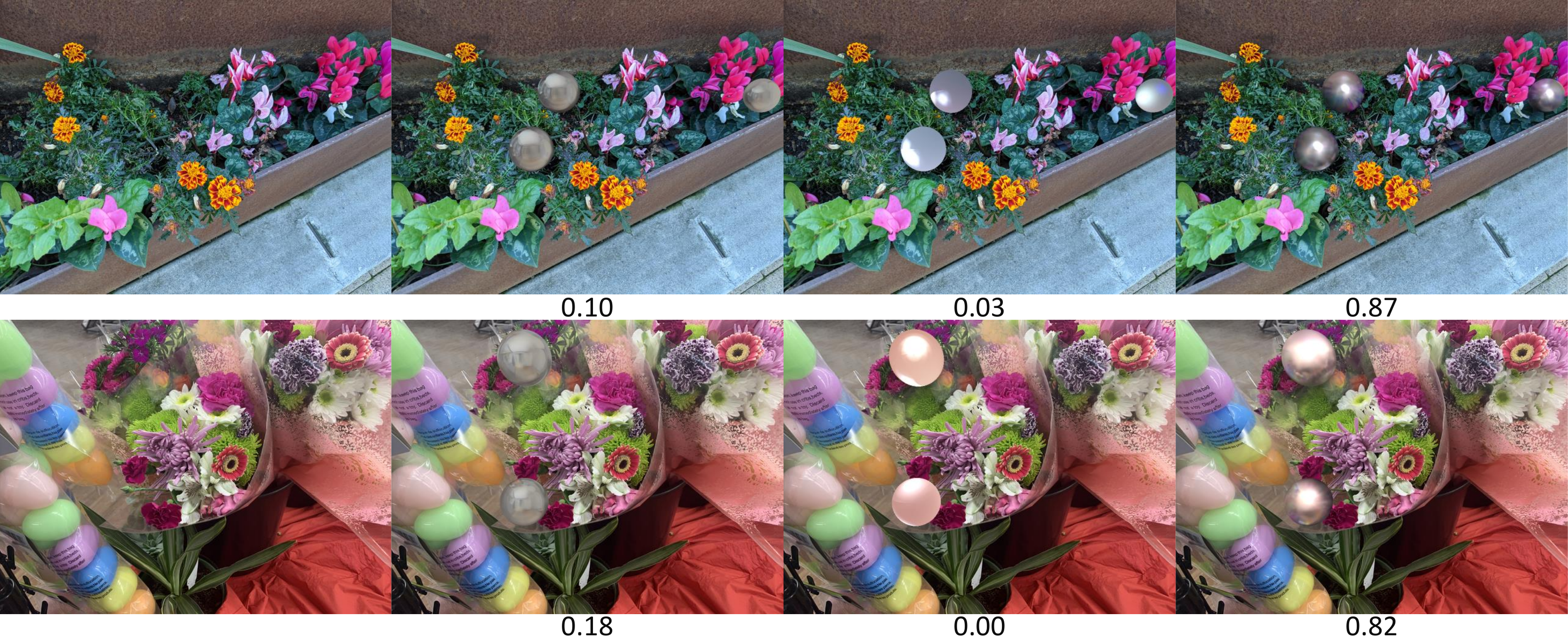}
   \caption{Additional chrome sphere insertion results on IBRNet dataset\cite{ibrnet}. The number under the image is the result of user study.}
   \label{fig:supp_oi_chrome1}
\end{figure*}

\begin{figure*}[ht]
  \centering
  \includegraphics[width=\linewidth]{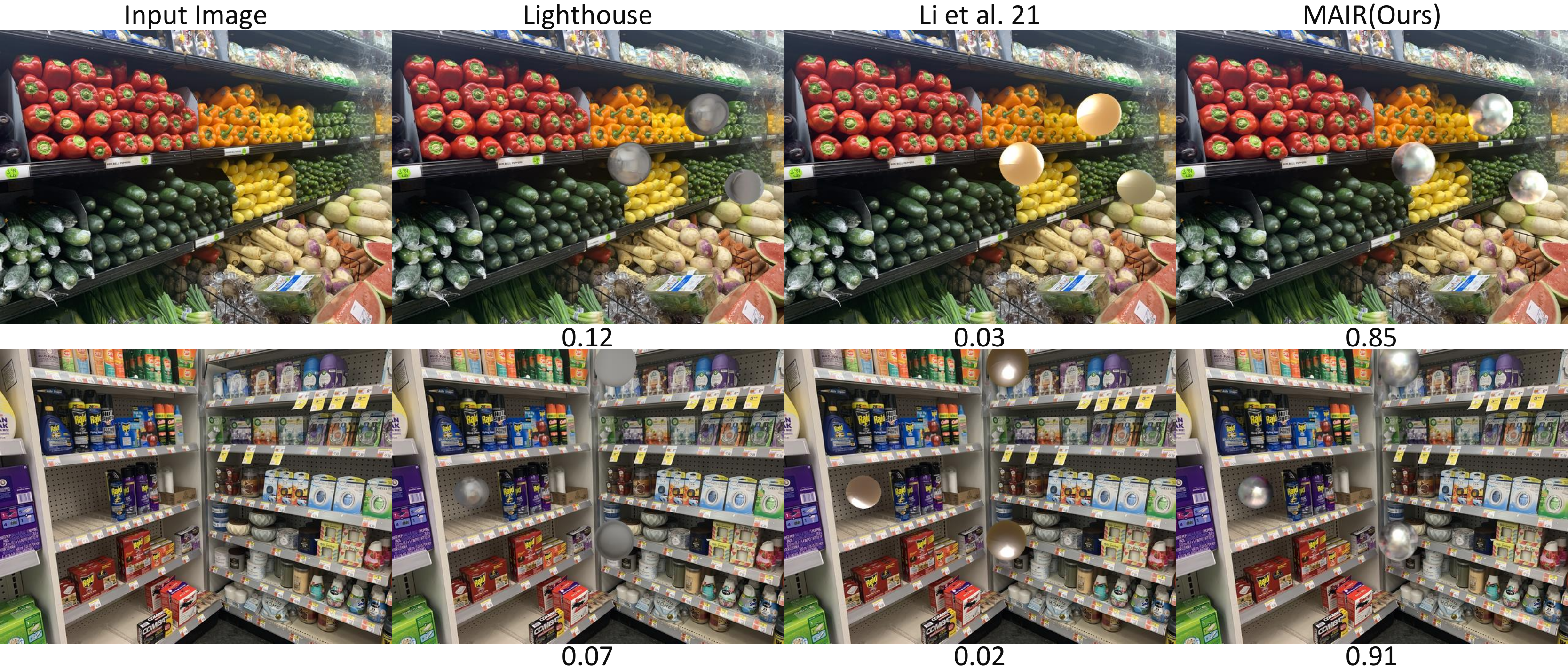}
   \caption{Additional chrome sphere insertion results on IBRNet dataset\cite{ibrnet}. The number under the image is the result of user study.} 
   \label{fig:supp_oi_chrome2}
\end{figure*}

\begin{figure*}[ht]
  \centering
  \includegraphics[width=\linewidth]{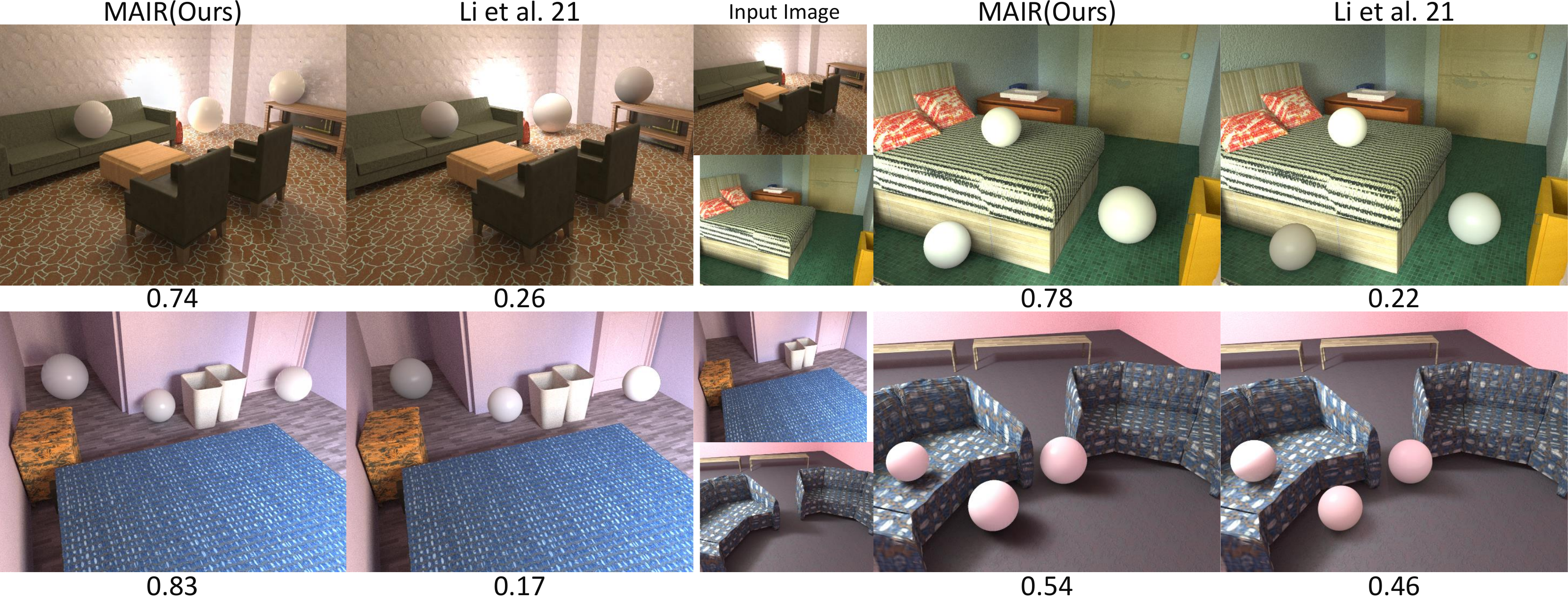}
  \includegraphics[width=\linewidth]{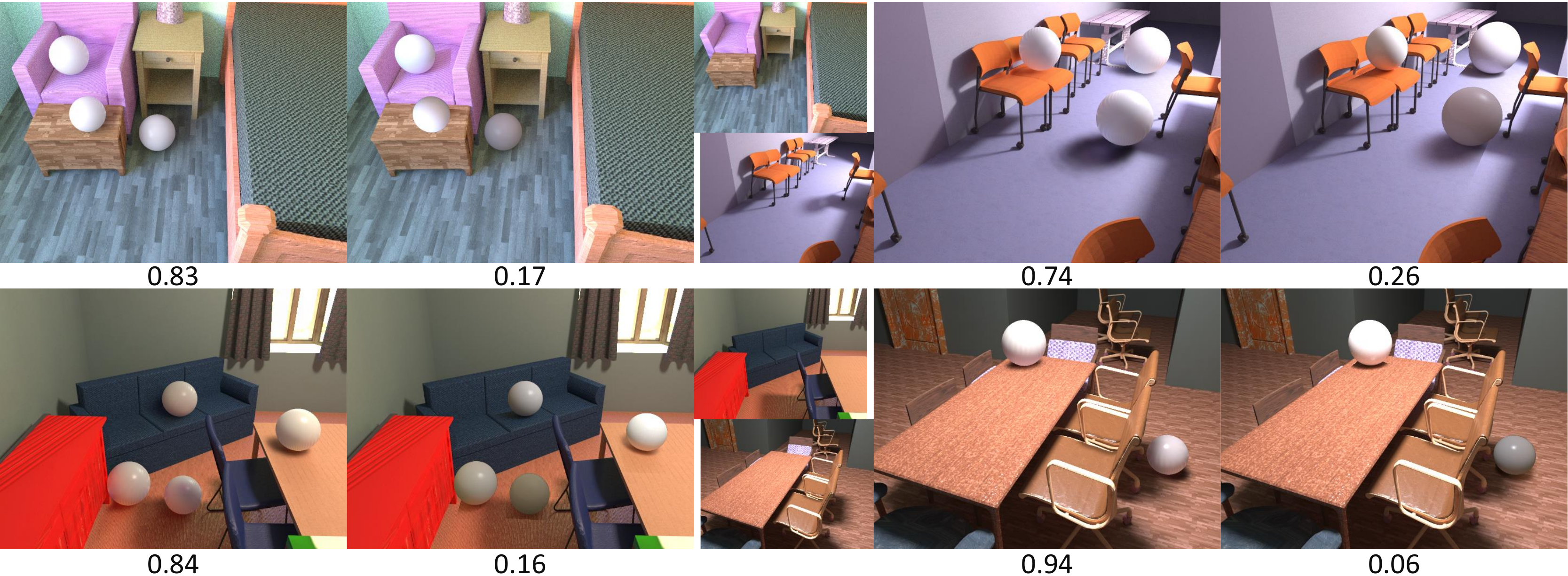}
   \caption{Additional white sphere insertion results on OpenRooms FF. The number under the image is the result of user study.} 
   \label{fig:supp_oi_indoor}
\end{figure*}

\begin{figure*}[ht]
  \centering
  \includegraphics[width=\linewidth]{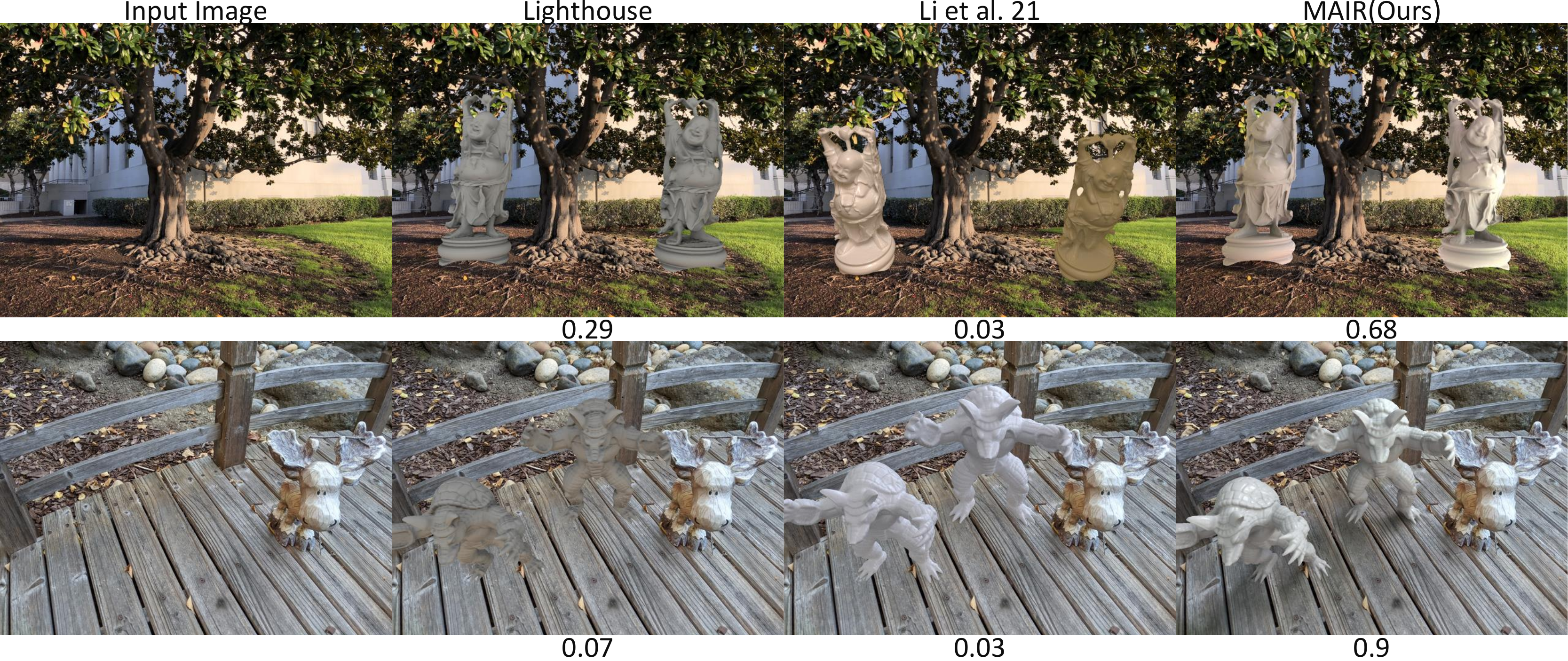}
  \includegraphics[width=\linewidth]{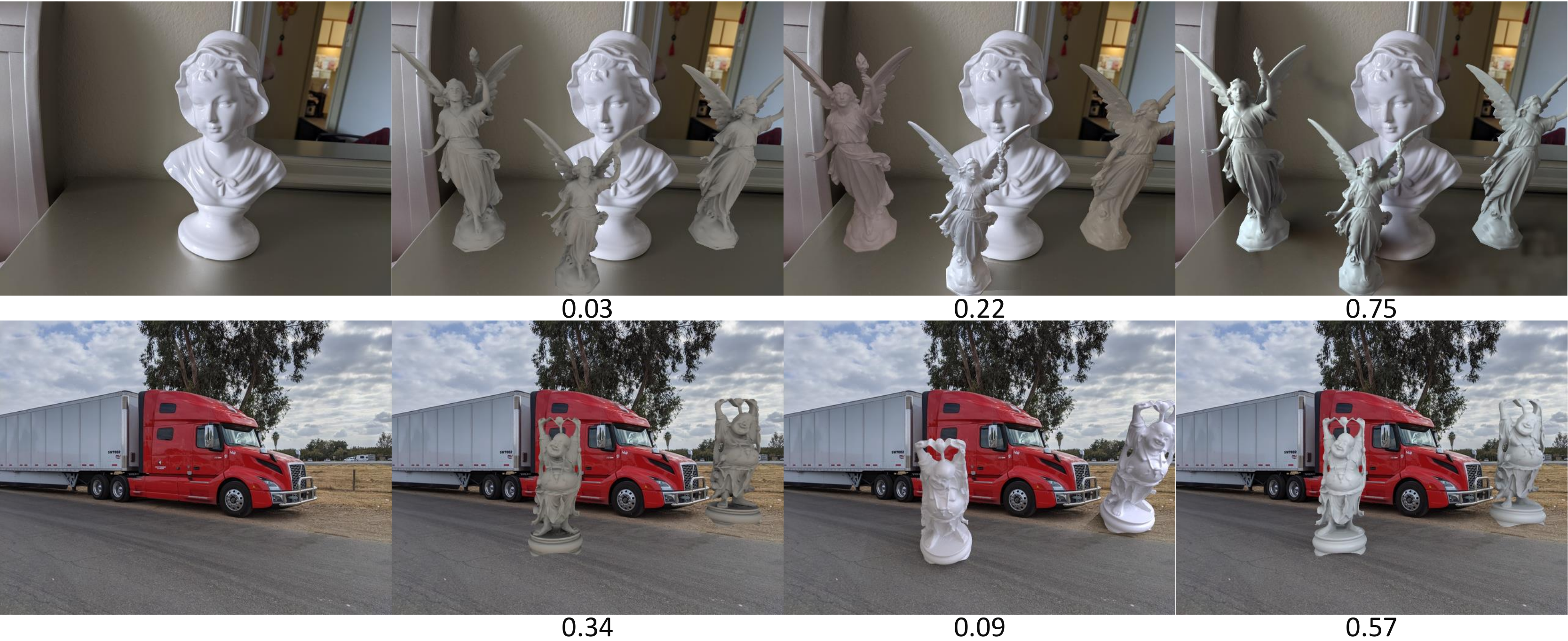}
  \includegraphics[width=\linewidth]{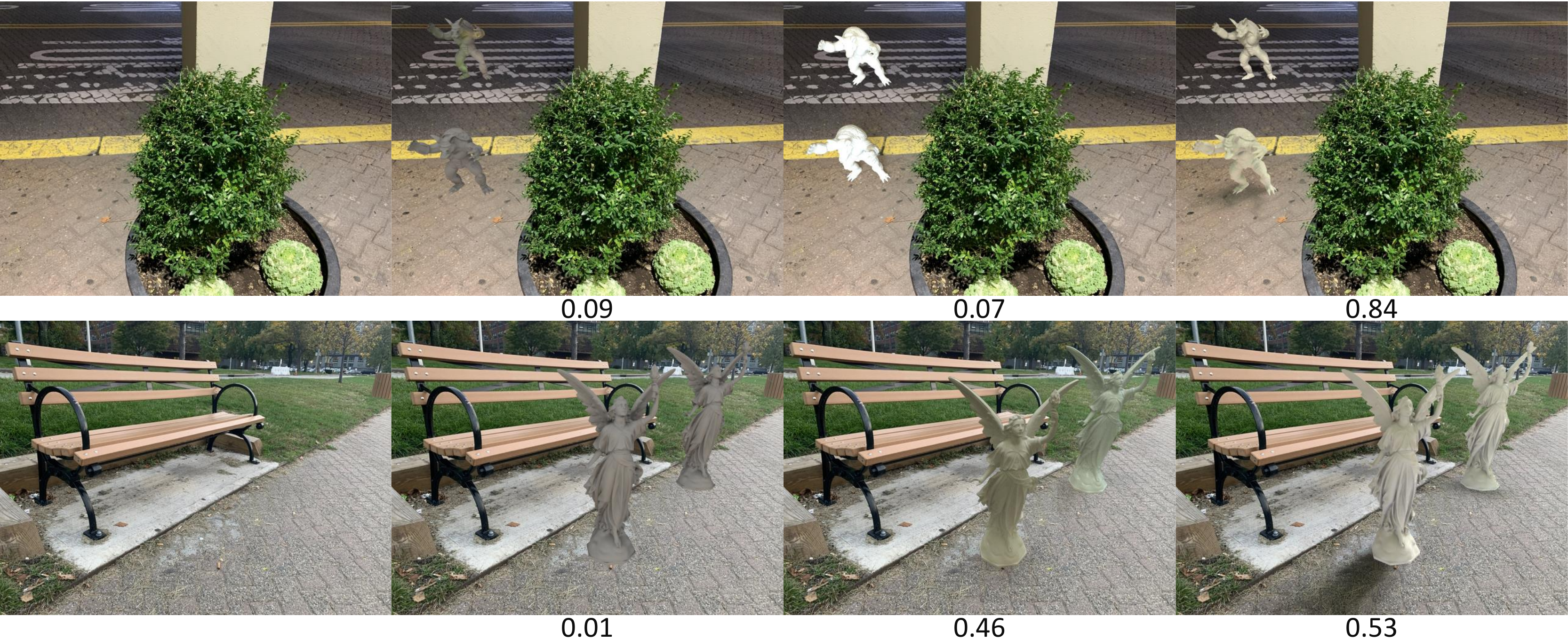}
   \caption{Additional virtual object~\cite{stanford} insertion results on IBRNet dataset\cite{ibrnet}. The number under the image is the result of user study.}
   \label{fig:supp_oi_real1}
\end{figure*}

\begin{figure*}[ht]
  \centering
  \includegraphics[width=\linewidth]{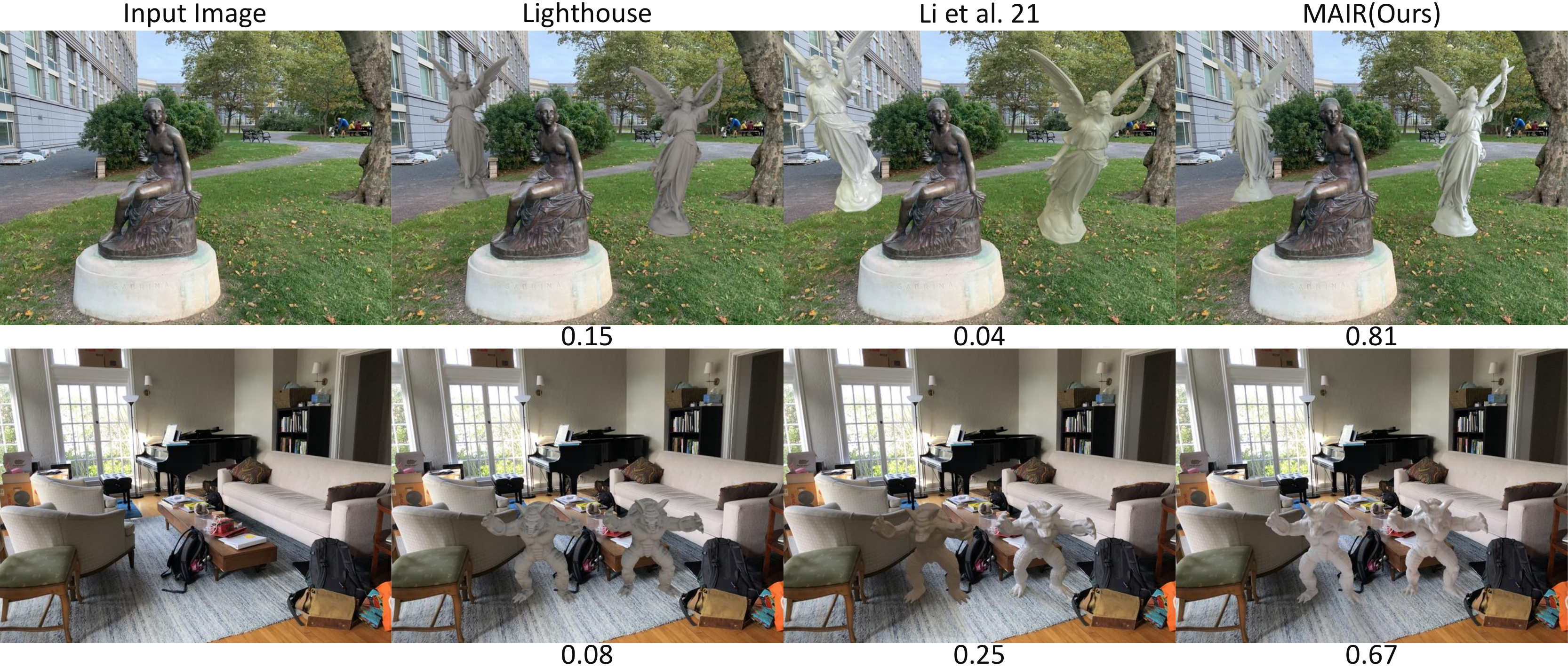}
  \includegraphics[width=\linewidth]{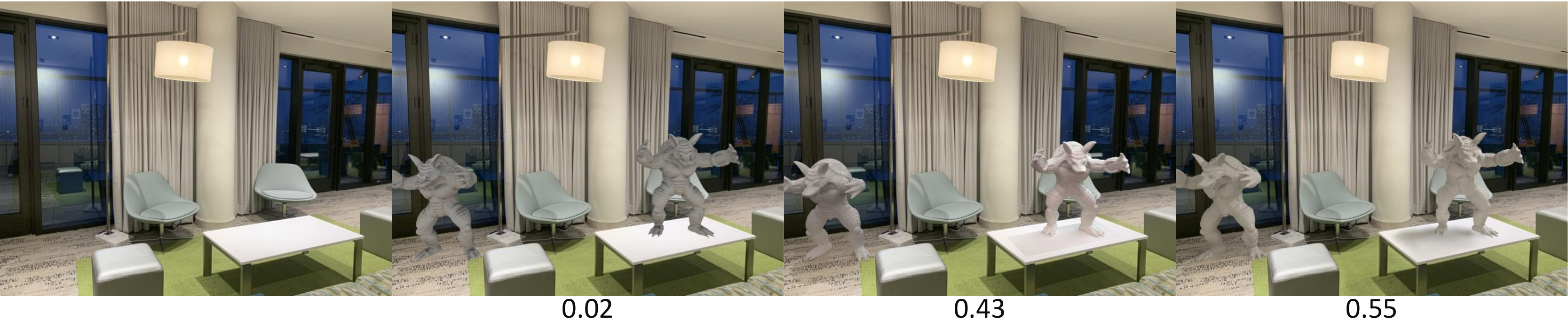}
   \caption{Additional virtual object~\cite{stanford} insertion results on IBRNet dataset\cite{ibrnet}. The number under the image is the result of user study.}
   \label{fig:supp_oi_real2}
\end{figure*}

We conducted a user study to evaluate the quality of object insertion from the three methods. Given a background image and an object of a particular material, users selected the most natural image among the three different results in a random order. 100 users evaluated 25 different scenes. Fig.~\ref{fig:supp_oi_chrome1}, \ref{fig:supp_oi_chrome2}, \ref{fig:supp_oi_indoor}, \ref{fig:supp_oi_real1}, and \ref{fig:supp_oi_real2} show all the scenes used in our user study. Our 3D lighting not only clearly expresses HDR lighting, but also fully reflects real-world scene geometry and material. This allowed the object to be realistically inserted into the scene, acquiring the highest score among the competing methods. 

We also provide additional object insertion results. In the accompanied video, the object can be located not only on the plane but also on any geometry, and the shadow of the object realistically appears to match the scene illumination.


{\small
\bibliographystyle{ieee_fullname}
\bibliography{egbib}
}

\end{document}